%% file: 0_main_gdm.tex
\documentclass[11pt, a4paper, logo, copyright]{googledeepmind}

\usepackage[authoryear, sort&compress, round]{natbib}
\usepackage{float}
\usepackage{subcaption}
\usepackage{makecell}
\usepackage{wrapfig}

\usepackage{listings}
\usepackage{xcolor}
\usepackage{microtype}

\usepackage{tikz}
\usetikzlibrary{shapes.geometric, arrows, positioning, shadows, calc}

\tikzstyle{recipe} = [rectangle, draw, fill=blue!20, text centered, rounded corners, minimum height=4em, text width=6em]
\tikzstyle{assets} = [rectangle, draw, fill=green!20, text centered, rounded corners, minimum height=4em, text width=6em]
\tikzstyle{area} = [rectangle, draw, fill=yellow!20, text centered, rounded corners, minimum height=4em, text width=6em]
\tikzstyle{line} = [draw, -latex']

\newcommand{\method}{\textsc{Proc4Gem}\:}
\renewcommand{\today}{}

\title{
\method:
 Foundation models for physical agency through procedural generation
}

\correspondingauthor{yixin@yixinlin.net}

\author[$\dagger$]{Yixin~Lin}
\author[$\dagger$]{Jan~Humplik}
\author[$\dagger$]{Sandy~H.~Huang}

\author[*]{Leonard~Hasenclever}
\author[*]{Francesco~Romano}
\author[*]{Stefano~Saliceti}
\author[*]{Daniel~Zheng}
\author[*]{Jose~Enrique~Chen}
\author[*]{Catarina~Barros}
\author[*]{Adrian~Collister}
\author[*]{Matt~Young}
\author[*]{Adil~Dostmohamed}
\author[*]{Ben~Moran}

\author[ \kern-0.2em]{Ken~Caluwaerts}
\author[ \kern-0.2em]{Marissa~Giustina}
\author[ \kern-0.2em]{Joss~Moore}
\author[ \kern-0.2em]{Kieran~Connell}

\author[$\ddagger$]{Francesco~Nori}
\author[$\ddagger$]{Nicolas~Heess}
\author[$\ddagger$]{Steven~Bohez}
\author[$\ddagger$]{Arunkumar~Byravan}

\affil[$\dagger$]{Co-first author}
\affil[*]{Core contributor}
\affil[$\ddagger$]{Co-last author}
\affil[ \kern-0.2em]{All work done at Google DeepMind.}

\begin{abstract}
In robot learning, it is common to either ignore the environment semantics, focusing on tasks like whole-body control which only require reasoning about robot-environment contacts, or conversely to ignore contact dynamics, focusing on grounding high-level movement in vision and language.
In this work, we show that advances in generative modeling, photorealistic rendering, and procedural generation allow us to tackle tasks requiring \textit{both}.
By generating contact-rich trajectories with accurate physics in semantically-diverse simulations, we can distill behaviors into large multimodal models that directly transfer to the real world: a system we call \method.
Specifically, we show that a foundation model, Gemini, fine-tuned on only simulation data, can be instructed in language to control a quadruped robot to push an object with its body to unseen targets in unseen real-world environments.
Our real-world results demonstrate the promise of using simulation to imbue foundation models with physical agency. Videos can be found at our website: \href{https://sites.google.com/view/proc4gem}{sites.google.com/view/proc4gem}.
\end{abstract}

\begin{document}

\maketitle

\section{Introduction}

\begin{figure*}[!ht]
    \centering
    \includegraphics[width=\textwidth]{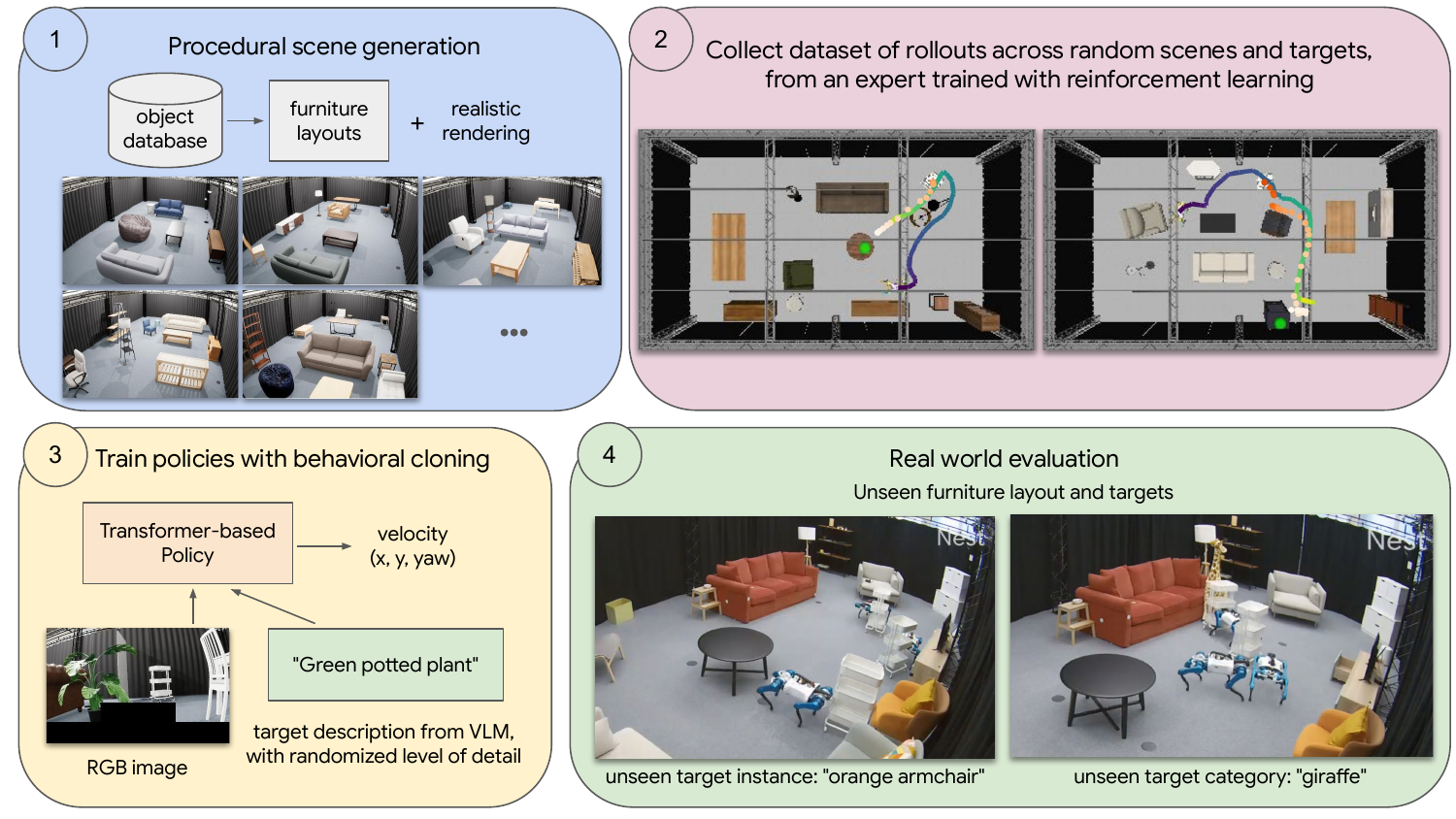}
    \caption{Method overview: (1) We use a dataset of furniture objects and random procedural generation to sample living room scenes for training (Section~\ref{sec:methods-sim}). (2) We use reinforcement learning to train a privileged trolley-pushing policy, and then collect rollouts from this policy while rendering high-resolution RGB images using Unity and generating captions for target objects using Gemini (Section~\ref{sec:methods-task-experts}). (3) We use behavior cloning to train transformer-based policies on this vision-language-action data (Section~\ref{sec:methods-distillation}). (4) We evaluate these policies in a held-out real-world scene (Section~\ref{sec:results}).}
    \label{fig:method_overview}
\end{figure*}

Recent hope of transferring the impressive capabilities in multimodal foundation models \citep{achiam2023gpt,team2023gemini} for robotics has been gated on the paucity of robot data relative to Internet-scale text and image data.
While there are initial signs of manual data collection being sufficient for some generality, for example by pooling together data across embodiments \citep{ebert2021bridge,padalkar2023open}, robotics as a data modality is still many orders of magnitude off from achieving comparable scale to the trillions of tokens routinely used to train frontier language models \citep{dubey2024llama}.

The combination of simulation and modern policy optimization techniques promises to address this by freeing data collection from real-world constraints, but key questions of an inevitable sim2real gap and task diversity prevent simulation from fully solving the problem and becoming an unlimited source of diverse, realistic data.
High-fidelity physics simulators like Mujoco \citep{todorov2012mujoco} and PyBullet \citep{coumans2021}, especially on accelerators \citep{makoviychuk2021isaac,freeman2021brax}, are highly effective at bridging this for specific tasks, leading to impressive feats of control in dexterous manipulation \citep{akkaya2019solving, rajeswaran2017learning, bauza2024demostart} and locomotion \citep{ouyang2024long,cheng2023legs,scholz2011cart,vasilopoulos2018sensor}.
However, they lack photorealism and semantic diversity and are thus not yet a solution for generating large-scale diverse data.

In contrast, embodied AI simulators like Thor \citep{kolve2017ai2}, Habitat \citep{puig2023habitat,szot2021habitat,savva2019habitat}, and Gibson \citep{xia2018gibson,xia2019gibson} achieve large-scale semantic diversity by trading off physical realism, especially through procedural generation \citep{deitke2022} or leveraging already-existing data \citep{chang2017matterport3d}.
Though they may enable some level of abstracted interactivity, this is mainly limited to tasks such as mobile pick-and-place.
There are early signs that the combination of realism and diversity can be effective for mobile manipulation in the real world when mixed with real data \citep{nasiriany2024robocasa}, but effective transfer of policies trained purely in simulation has not yet been shown.

This landscape seems to impose a dichotomy between tasks which are either simplified from a physics perspective or contain limited semantic diversity.
Can we instead tackle the more general problem of contact-rich robotic control in semantically diverse environments?

As a step towards this goal, we consider the task of controlling a quadrupedal robot to push a trolley towards diverse objects specified using natural language, in rooms with varied layouts.
This is a natural choice because it requires fusing high-level semantic generalization capabilities with low-level physical reasoning: it subsumes both the problems of semantic navigation and whole-body manipulation of an object.
It also precludes a basic modular solution: while the type of abstracted mobile manipulation can often be solved with the sequential application of separate navigation/manipulation policies, our chosen task and embodiment pushes us to develop an end-to-end solution that combines both physical and semantic reasoning in the same policy.

Our main contribution is therefore a multi-step system, called \method, for tackling whole-body control grounded in semantic understanding, described in Figure~\ref{fig:method_overview}.
We find that this system can turn Gemini, a large multimodal model, into an effective policy for language-conditioned whole-body control purely through fine-tuning on simulated data.
We further demonstrate that this Gemini policy outperforms a strong baseline, SPOC~\citep{ehsani2024spoc}, trained on the same data, in terms of real-world performance and generalization.
Our hope is that this work demonstrates the effectiveness of bringing photorealistic rendering, high-fidelity physics, and diverse procedural generation together as an effective method for synthetically generating large-scale robot data for policies which directly transfer to the real world.

\input{methods}

\input{results}
\input{related_work}

\section{Conclusion}

In this work, we demonstrate a system, \method, which combines physically-realistic simulations with high semantic diversity to generated large-scale synthetic robot action data for a task requiring physical reasoning.
We show that large multimodal models like Gemini, fine-tuned solely on this simulation data, can directly accomplish this task in the real world and demonstrate strong generalization.

The fundamental longer-term promise of highly diverse yet accurate simulations, as an instance of synthetic data generation, is the ability to directly \textit{convert compute into data}.
If this promise is fully realized, these methods could supplement or even entirely replace real-world data collection and enable a path to generating the magnitude of data necessary for scaling robotic foundation models.

Fruitful future work include extending to tasks able to leverage the already-massive context windows of multimodal models like Gemini, pushing more deeply on augmenting the simulation by exploiting modern image/video generation models, and moving beyond behavior-cloning to direct reinforcement learning of these large models within simulations.

\section{Acknowledgments}

We would like to thank Mohamed Amtouti for significant help in running hardware experiments, and Oliver Groth for valuable feedback on the paper.

\bibliography{1_bib}

\clearpage
\input{appendix}

\end{document}

%% file: methods.tex
\section{Methods}

\subsection{Procedural scene generation and simulation}\label{sec:methods-sim}

As described in Figure~\ref{fig:method_overview}, the first step of our system is a protocol for sampling realistic scenes.
We follow recent work on procedural generation \citep{deitke2022, li2023behavior}, and use a VLM-captioned dataset of assets along with a heuristic hierarchical sampler for generating living-room like scenes.
In contrast to previous works, we significantly increase the realism of our simulations in order to improve sim2real transfer.
To enable realistic dynamics, including detailed object interactions, we simulate the scenes in MuJoCo~\citep{todorov2012mujoco}.
We furthermore render the scenes in Unity to provide photo-realistic image observations.

\textbf{Asset data}
The asset dataset contains thousands of assets, that are primarily large pieces of furniture, e.g., sofas, chairs, dining tables, and coffee tables.
We use Gemini to produce five natural language descriptions per asset, with increasing level of detail.
These descriptions are used as language commands for our agents. See Appendix Fig.~\ref{fig:description_examples} for examples of these descriptions.

\begin{figure*}
    \centering
    \includegraphics[width=\textwidth]{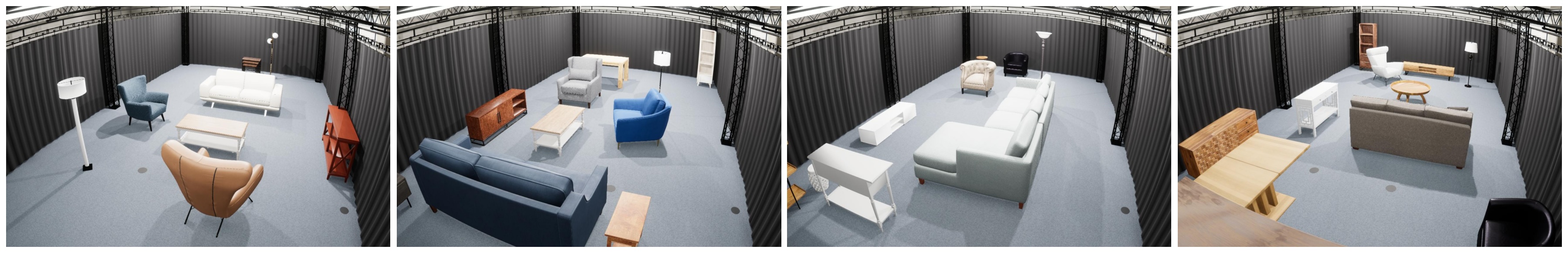}
    \caption{Examples of procedurally-generated living room scenes. We use a hierarchical placement recipe to place assets in semantically-meaningful configurations.}
    \label{fig:living_room_samples}
\end{figure*}

\textbf{Asset placement}
We procedurally generate indoor, living-room like spaces by sampling and placing assets in semantically-meaningful configurations.
We use a heuristic, hierarchical approach based on a placement recipe.
Starting from the overall room, we first sample (possibly nested) areas (e.g., a dining area) and then instances of relevant asset categories within each area.
Fig.~\ref{fig:living_room_samples} shows a number of sampled living room scenes following this recipe. Note that for this work we do not place any assets on top of other assets, despite e.g. a TV and cushions being present in the real-world evaluation setup.

\begin{figure}[!hb]
    \centering
    \begin{subfigure}[b]{0.49\columnwidth}
        \centering
        \includegraphics[width=\textwidth]{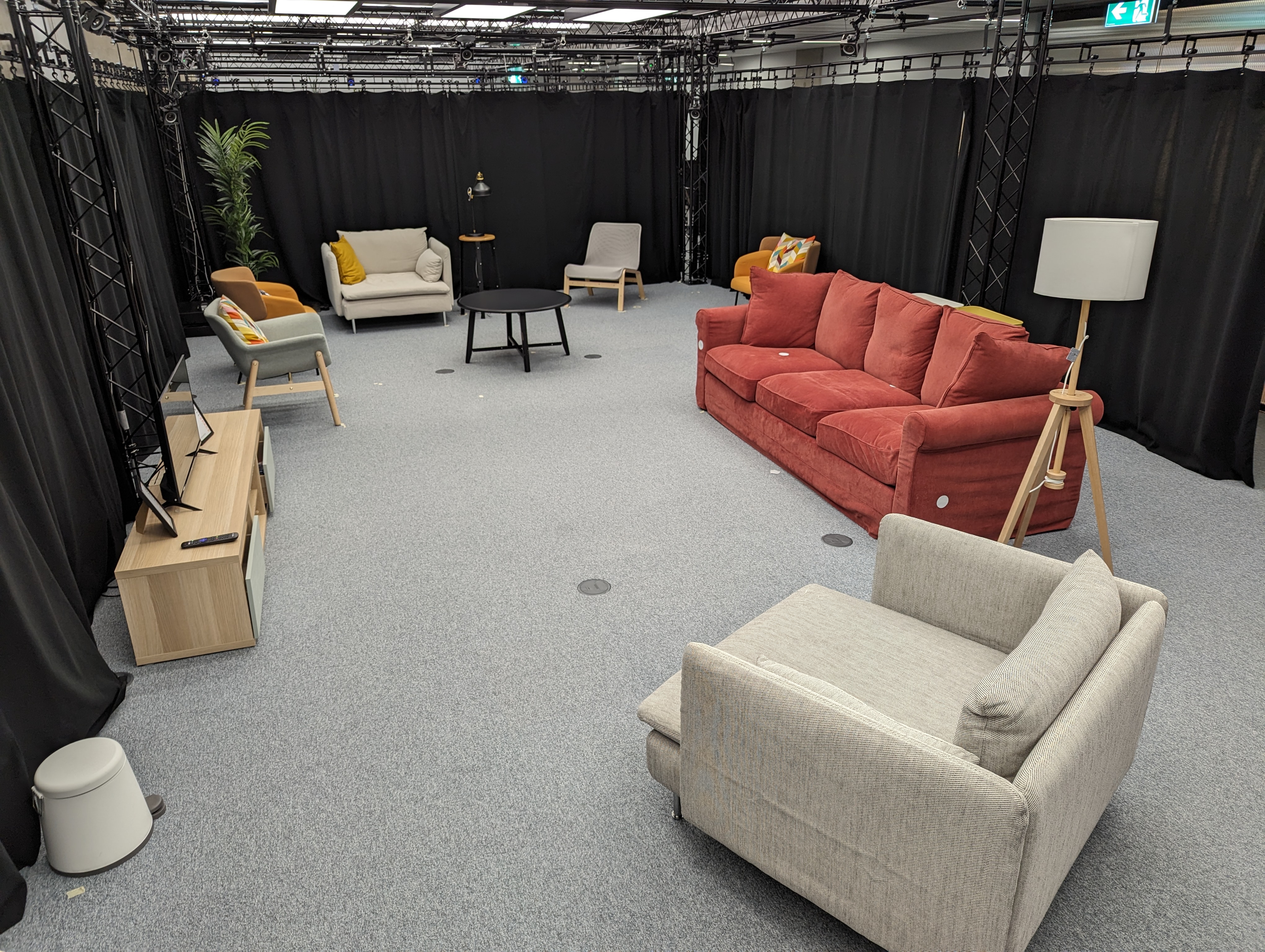}
        \caption{Real world.}
    \end{subfigure}
    \begin{subfigure}[b]{0.49\columnwidth}
        \centering
        \includegraphics[width=\textwidth]{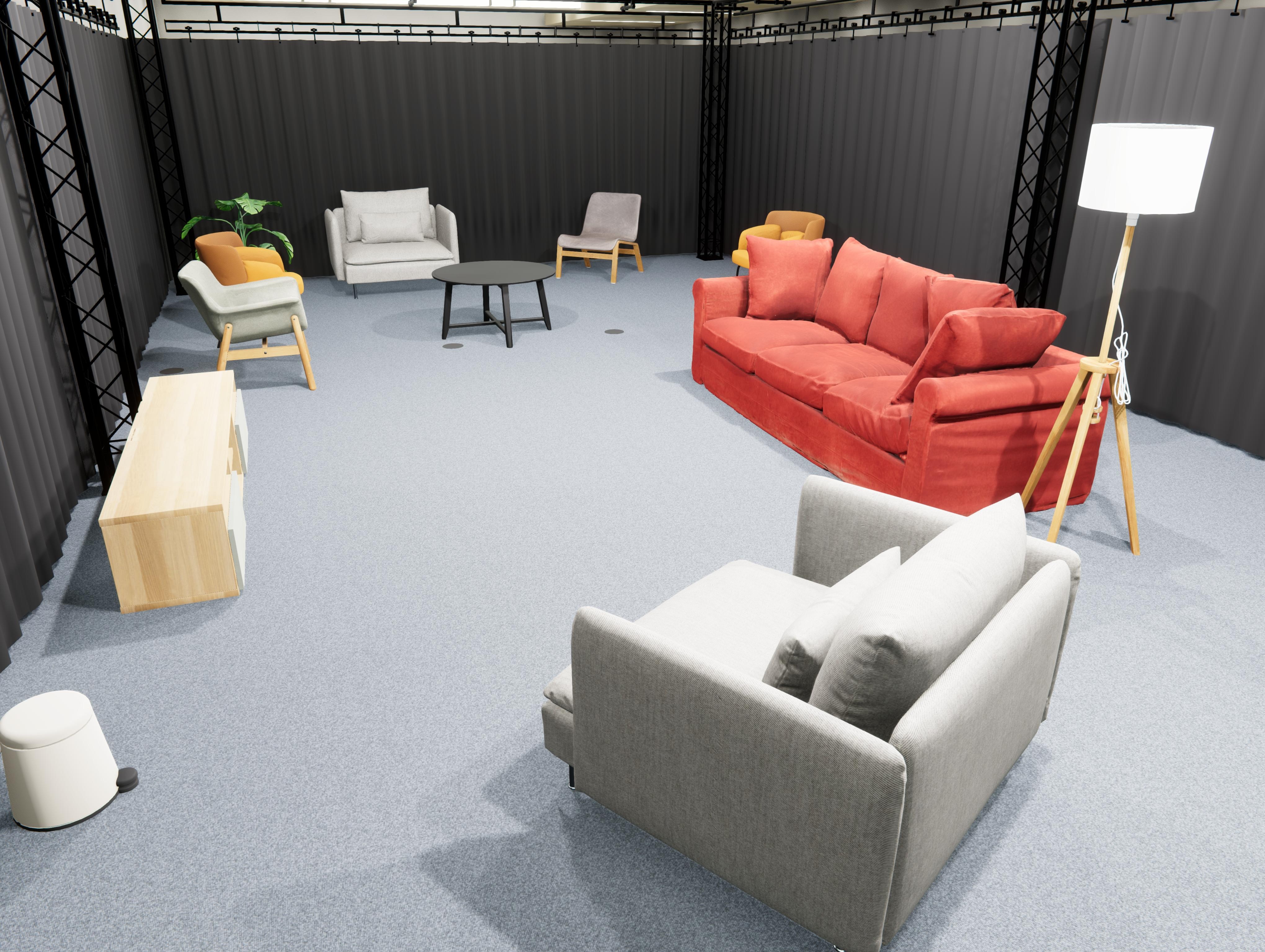}
        \caption{Unity rendering.}
    \end{subfigure}
    \caption{Comparison between the real-world setup and the equivalent simulated scene with Unity rendering. This scene is not used for training data generation, only for evaluation.}
    \label{fig:unity-sim-real}
\end{figure}

\textbf{Realistic rendering}
Deploying robot controllers that have been trained with simulated RGB vision in reality has historically been challenging due to the visual domain gap between images rendered from simulations and captured with real cameras.
While recent techniques such as NeRF2Real~\citep{byravan2023nerf2real, deitke2023phone2proc, shafiullah2022clip} have opened up new opportunities, they are typically limited to a small number of (mostly static) scenes which are often directly the scenes where the robot is to be deployed.
Modern game engines, on the other hand, provide a myriad of settings to enable realistic rendering of scenes.
In our setup we offload the rendering of high-resolution and high-fidelity RGB images to Unity~\citep{unity}.
With GPU-support and efficient multi-view rendering, we observe little to no overhead while generating highly-realistic images, as can be seen in Fig.~\ref{fig:unity-sim-real}.

\subsection{Task and expert policies in simulation}\label{sec:methods-task-experts}

Our goal is to train a policy which can push a white trolley to any object within any random scene layout. Target objects are sampled uniformly from all objects present in the scene. An episode is successful if the robot succeeds in pushing the trolley to the goal in under 30 seconds without flipping over (60 seconds during evaluation).

\subsubsection{Robot platform and low-level controller}\label{section:platform}

\begin{wrapfigure}{r}{0.4\textwidth}
    \centering
    \includegraphics[width=0.38\textwidth]{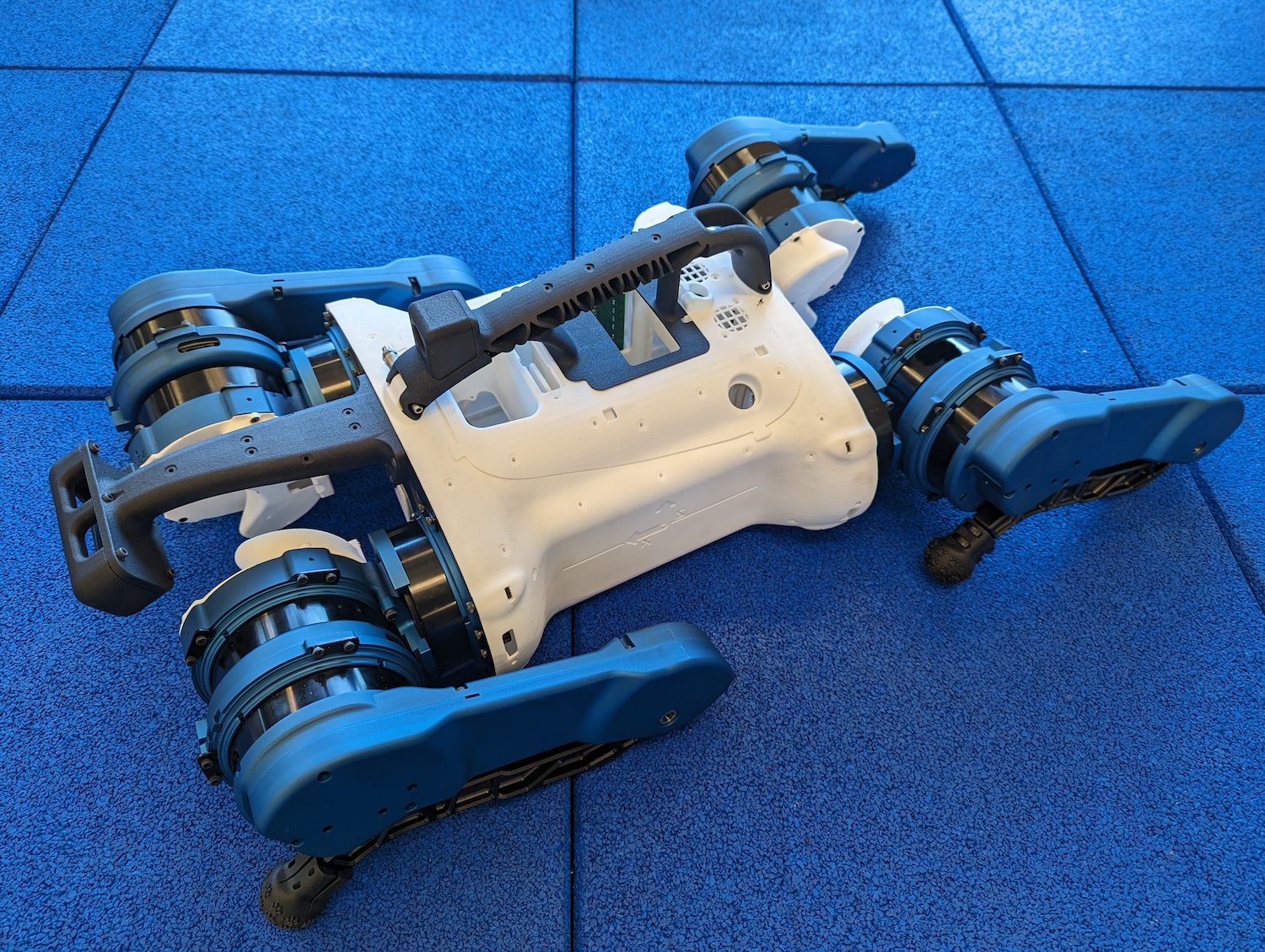}
    \caption{Barkour robot.}
    \label{fig:barkour_robot}
\end{wrapfigure}

We use the Barkour robot\footnote{\href{https://github.com/google-deepmind/barkour_robot}{github.com/google-deepmind/barkour\_robot}} \citep{caluwaerts2023barkour}, depicted in Fig.~\ref{fig:barkour_robot}, as the embodiment for our experiments. This robot has two onboard cameras. Due to the placement of the camera on its back, a part of the robot is always in view, so we apply a simple black mask to hide the self-occlusion.
The robot's MuJoCo model is publicly available in the MuJoCo Menagerie~\citep{menagerie2022github}.

Our policies output a quantized 3D planar velocity command for this robot at a frequency of 2 Hz. The command dimensions correspond to forward-backward movement, right-left side-stepping, and yaw turning. The low-level controller which maps these velocity commands to robot motion was trained using on-policy reinforcement learning and massively parallel TPU-accelerated MuJoCo XLA (MJX) simulation.\footnote{\href{https://mujoco.readthedocs.io/en/stable/mjx.html}{mujoco.readthedocs.io/en/stable/mjx.html}}

The low control frequency of 2 Hz is a practical choice based on model inference speed, which does constrain the difficulty of the control problem. However, we note that despite this, the model still needs to reason about its physical state (e.g. momentum) and the effect of its actions upon the world in order to properly manipulate the trolley with its body; furthermore, it needs to reactively recover from the inevitable imprecision of the action space.

\subsubsection{Training privileged experts}

In simulation, we first use model-free off-policy RL to train an expert policy that leverages privileged state information, and then we distill this into a student policy that only relies on high-resolution images and language instructions. Because it uses privileged information, the expert has a smaller policy network compared to the student's, and can be trained more efficiently and without expensive rendering. To robustify the expert policy, we add domain randomization across episodes.

Given this expert, we collect hundreds of thousands of successful episodes across procedurally generated scenes. Each episode contains textual descriptions of the target object sampled in that episode, images from both onboard cameras, and robot actions. We split episodes into trajectories of length eight, which forms our dataset for training the student.

\subsection{Distillation and deployment}\label{sec:methods-distillation}

We fine-tune Gemini \citep{reid2024gemini,team2023gemini}, a general-purpose large multimodal model, using the trajectory data described above with next-token prediction loss (i.e., behavioral cloning).

Due to constraints of deploying large multimodal models, we engineer a distributed, asynchronous hierarchical control system; see Appendix Section~\ref{appendix:hardware-deployment} for details. In particular, the robot hardware remotely queries the fine-tuned Gemini model, which induces meaningful network latency and jitter.
To deal with this, the most recent action returned by Gemini is cached and only updated to the low-level controller at regular intervals (in this case, at 2Hz).

%% file: results.tex
\section{Results}\label{sec:results}

We evaluated both the fine-tuned Gemini policy and a state-of-the-art baseline policy in simulation and the real world. Both models recover a large portion of the privileged expert's performance, and demonstrate semantic generalization to different objects, multi-lingual capabilities in language conditioning, and scene generalization to unseen layouts.
While both policies are trained with only simulated robot data, they directly transfer to the real world due to realistic rendering and physics domain randomization; the Gemini policy additionally generalizes to more difficult instances of the task and a novel target object category.

\subsection{Baseline training}

We reimplement the SPOC transformer-based architecture from \citet{ehsani2023imitating} as our state-of-the-art baseline. This architecture consists of pre-trained and frozen SigLIP text/image encoders \citep{zhai2023sigmoid}, a goal-conditioned transformer encoder, and a transformer decoder. The model takes in the same inputs as the Gemini policy, but with a separate image for each camera, resized to 224 x 224. We use a context length of $8$, to match that of the Gemini policy. 
We train the baseline on four epochs of data to obtain convergence; further training leads to mild declines in performance in the simulated environment, due to overfitting.

\subsection{Simulation evaluations}

\begin{figure}
    \begin{subfigure}[b]{0.54\textwidth}
        \centering
        \includegraphics[width=0.7\textwidth]{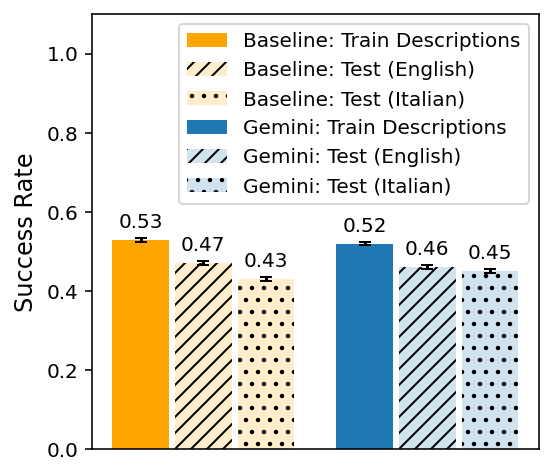}
        \caption{Evaluation of scene and language generalization in procedurally-generated simulated scenes, with 10,000 trials per setting. The error bars show standard error. The privileged RL expert achieves 68.9\% success.}
        \label{fig:results-sim-procgen-scene}
    \end{subfigure}
    \hfill
    \begin{subfigure}[b]{0.42\textwidth}
        \centering
        \includegraphics[width=0.9\textwidth]{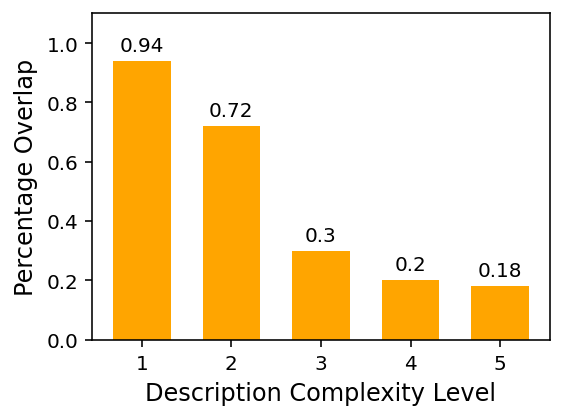}
        \caption{Overlap between train and test English descriptions at the different complexity levels. For examples of the descriptions at each level, please refer to Appendix Fig.~\ref{fig:description_examples}.
        }
        \label{fig:train-test-description-overlap}
    \end{subfigure}
    \caption{Simulation results in procedurally-generated scenes.}
\end{figure}

During evaluation, the policy is given 60 seconds to push the trolley to the target object.
The textual description of the target object is sampled from the five descriptions of increasing verbosity.
We evaluate policies in two settings in simulation: 1) in unseen procedurally-generated scenes, drawn from the same random distribution as the training data; and 2) in a replica of the real-world test living room, constructed by 3D-scanning the furniture and building a simulation which approximates their relative positions (Fig.~\ref{fig:unity-sim-real}).
In the procedurally-generated scenes, we investigate linguistic generalization capability by creating test sets of asset descriptions.
These are not necessarily fully disjoint from the training descriptions, and in fact at the lower description verbosity levels there is almost guaranteed to be overlap, as seen in Fig.~\ref{fig:train-test-description-overlap}. We also generate a set of descriptions in a different language (Italian) in order to test multilingual capabilities.

\begin{table}
    \small
    \centering
    \begin{tabular}{|l | c c|}
        \hline
            & \makecell{Trained without \\ 3D-scanned \\ assets} & \makecell{Trained with \\ 3D-scanned \\ assets} \\
            \hline
            Baseline & 59.6\% $\pm$ 0.49\% & 62.1\% $\pm$ 0.49\% \\
            Gemini & 53.0\% $\pm$ 0.50\% & 70.0\% $\pm$ 0.46\% \\
        \hline
    \end{tabular}
    \caption{Fixed simulation scene results, with mean and standard error over 10,000 trials. The privileged RL expert achieves 85.4\% success.
    }
    \label{tab:fixed-scene-results}
\end{table}

\subsection{Real-world hardware evaluations}

In the real world, we evaluate our policies trained without 3D-scanned assets in the mock living room scene.
We select three targets within the living room and three relative trolley/target/robot initial poses representing varying difficulty levels, and run 10 evaluation trials for each. Fig.~\ref{fig:results-hardware-with-initialization}b-d shows the difficulty levels for one of the targets; see Appendix~Table~\ref{tab:hardware-eval-positions} for all initial positions.
Each episode lasts up to 60 seconds and we terminate the episode on success, irrecoverable failure, or timeout.

While these real-world evaluations test out-of-distribution objects, they are drawn from a similar distribution as the training data; we go further and investigate whether the policies are robust to extremely out of distribution targets.
In particular, we choose a large toy giraffe with height 1.5 meters and the text description ``Giraffe,'' and run 10 trials on the medium difficulty setting.
We also perform ad-hoc tests of whether the Gemini policy has further semantic generalization capabilities, for example to humans or other robots, and whether the policies are robust to significantly different physical parameters, by adding 10 kilograms of weight to the trolley.

\begin{figure*}
    \begin{subfigure}[b]{1.0\textwidth}
        \centering
        \includegraphics[width=0.9\textwidth]{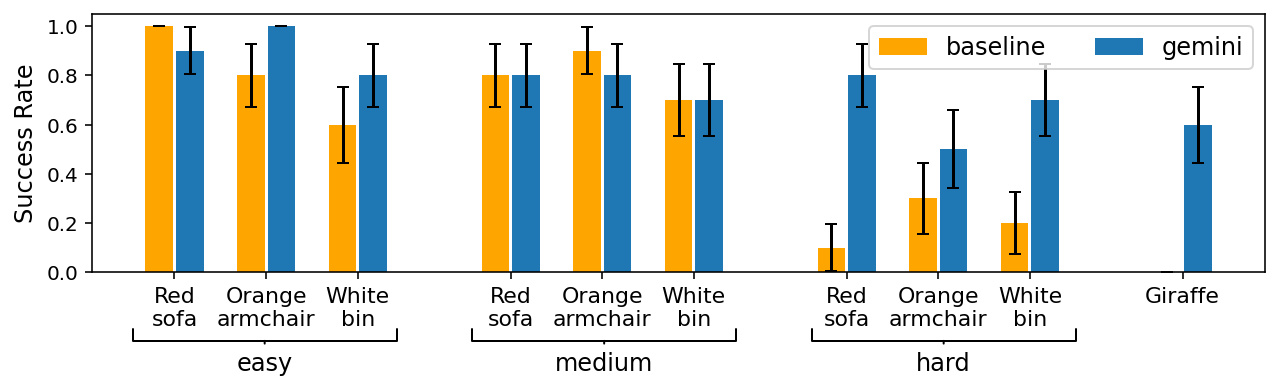}
        \caption{Full hardware experiment results, with ten trials per setting. Error bars indicate standard error.}
        \vspace{1em}
    \end{subfigure}
    \begin{subfigure}[b]{0.31\textwidth}
        \centering
        \includegraphics[width=\textwidth]{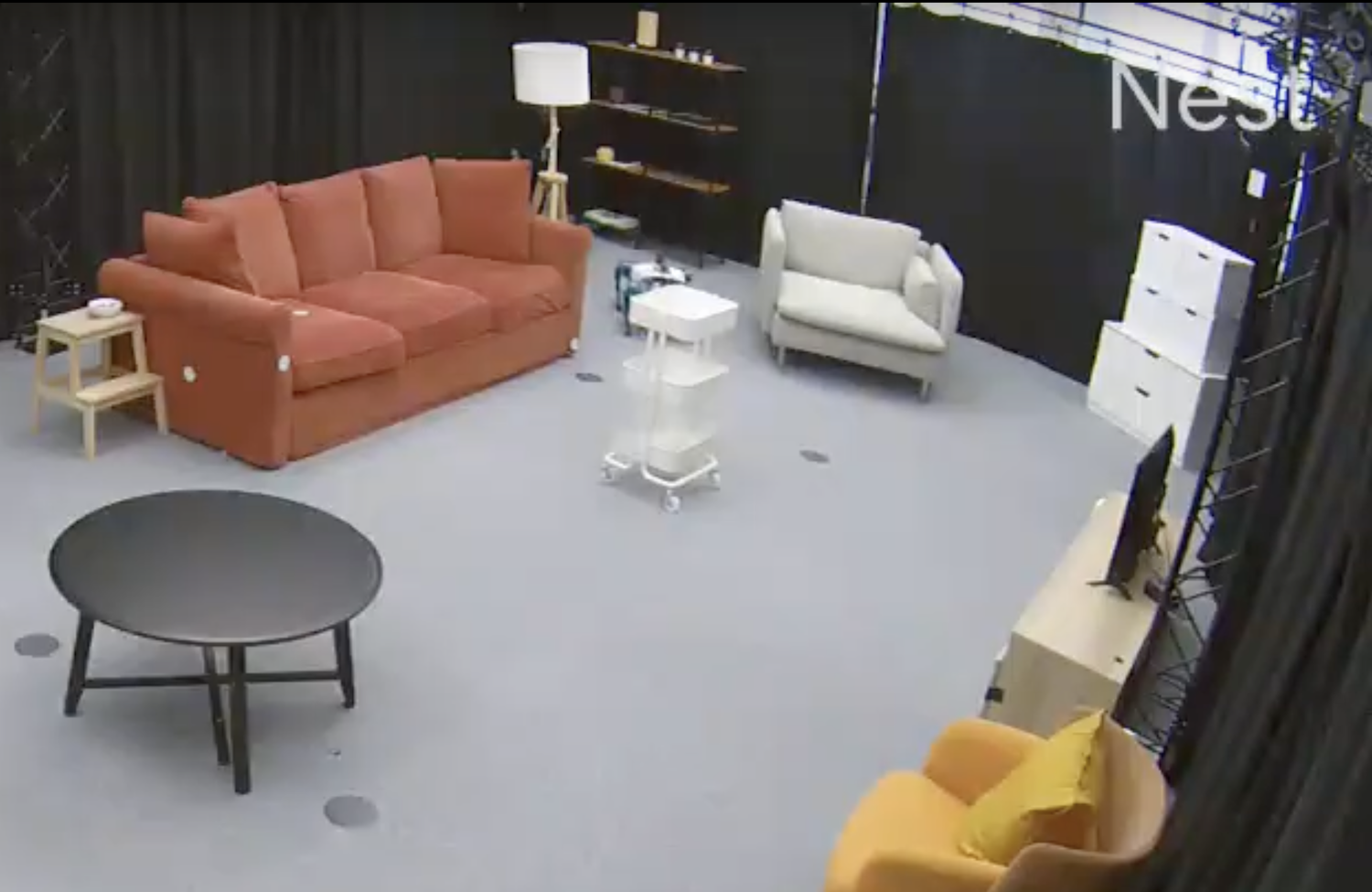}
        \caption{Orange armchair, easy.}
    \end{subfigure}
    \hspace{0.5em}
    \begin{subfigure}[b]{0.31\textwidth}
        \centering
        \includegraphics[width=\textwidth]{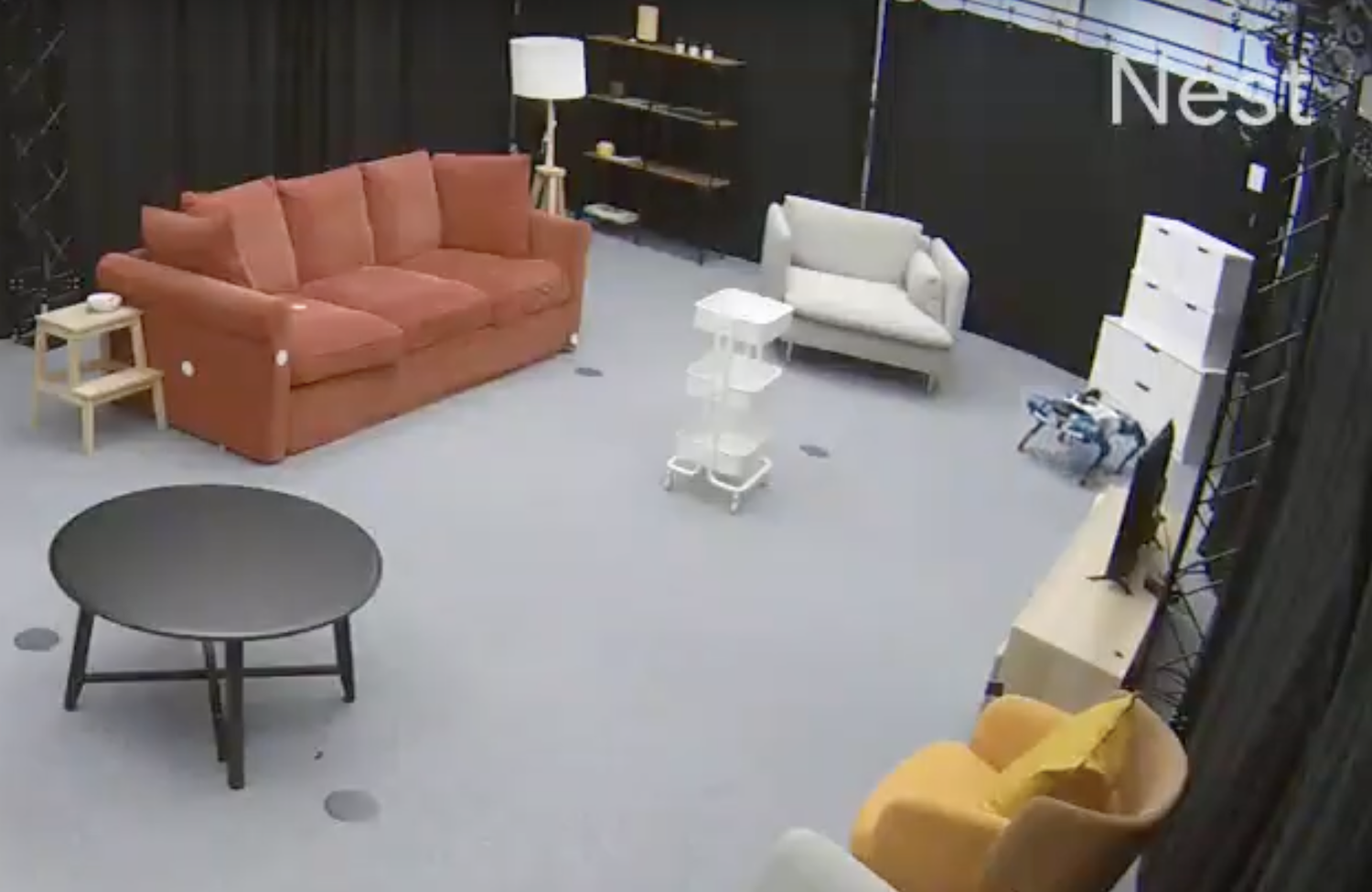}
        \caption{Orange armchair, medium.}
    \end{subfigure}
    \hspace{0.5em}
    \begin{subfigure}[b]{0.31\textwidth}
        \centering
        \includegraphics[width=\textwidth]{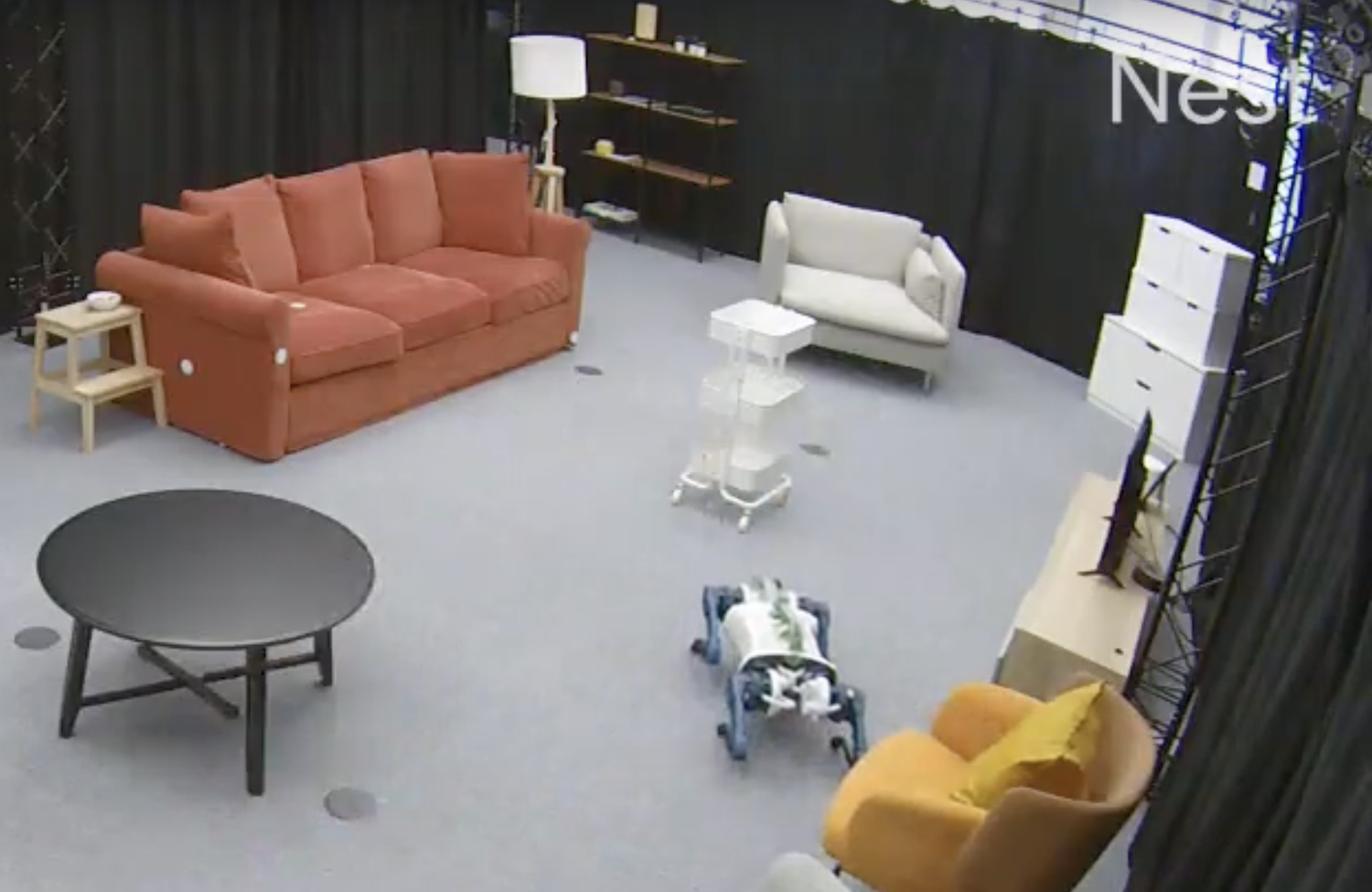}
        \caption{Orange armchair, hard.}
    \end{subfigure}
    \caption{Hardware experimental setup for a single target. In the \textit{easy} setting, the robot faces both the trolley and the target; in the \textit{medium} setting, the robot faces the trolley but not the target; and in the \textit{hard} setting, the robot faces the trolley while the target is behind it, so it needs to explore in order to find the target. See Appendix~Table~\ref{tab:hardware-eval-positions} for all evaluation initial positions.}
    \label{fig:results-hardware-with-initialization}
\end{figure*}

\subsection{Analysis}

\begin{figure*}
    \begin{subfigure}[b]{0.31\textwidth}
        \centering
        \includegraphics[width=\textwidth,height=0.524\textwidth]{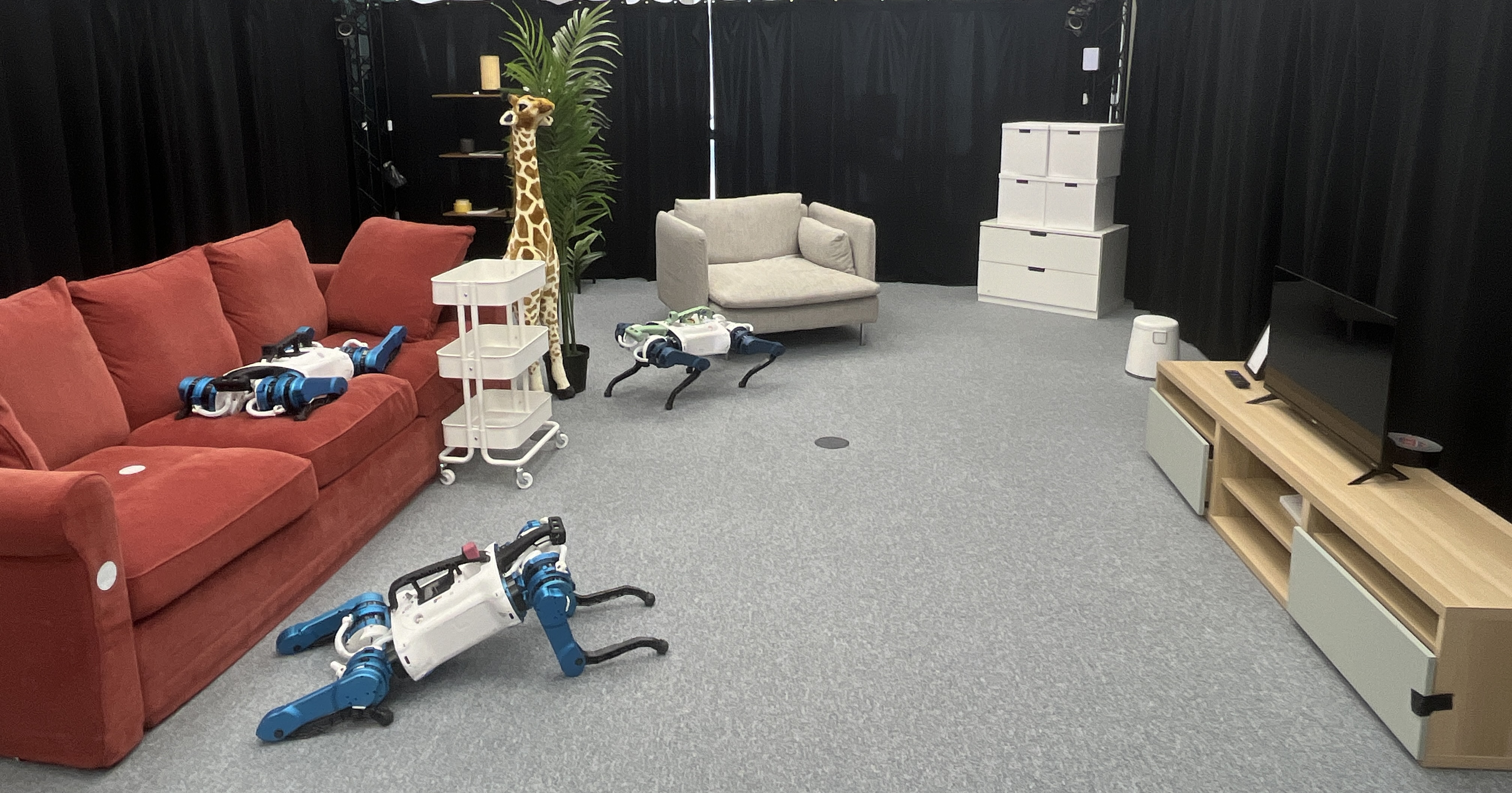}
        \caption{Robots and giraffe as targets in the scene.}
    \end{subfigure}
    \hspace{0.5em}
    \begin{subfigure}[b]{0.31\textwidth}
        \centering
        \includegraphics[width=\textwidth,height=0.524\textwidth]{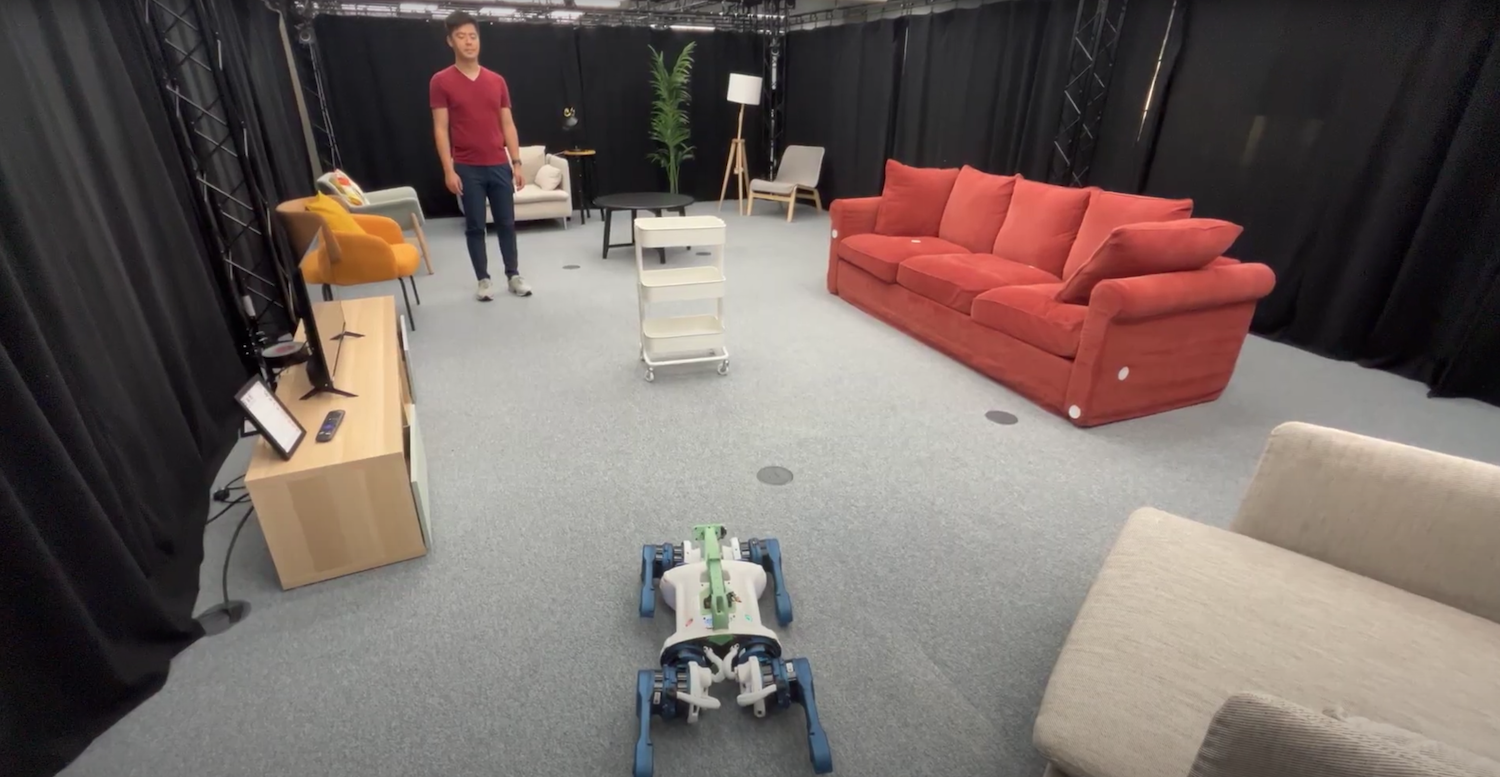}
        \caption{``Person with red shirt and blue pants'' as target.}
    \end{subfigure}
    \hspace{0.5em}
    \begin{subfigure}[b]{0.31\textwidth}
        \centering
        \includegraphics[width=\textwidth,height=0.524\textwidth]{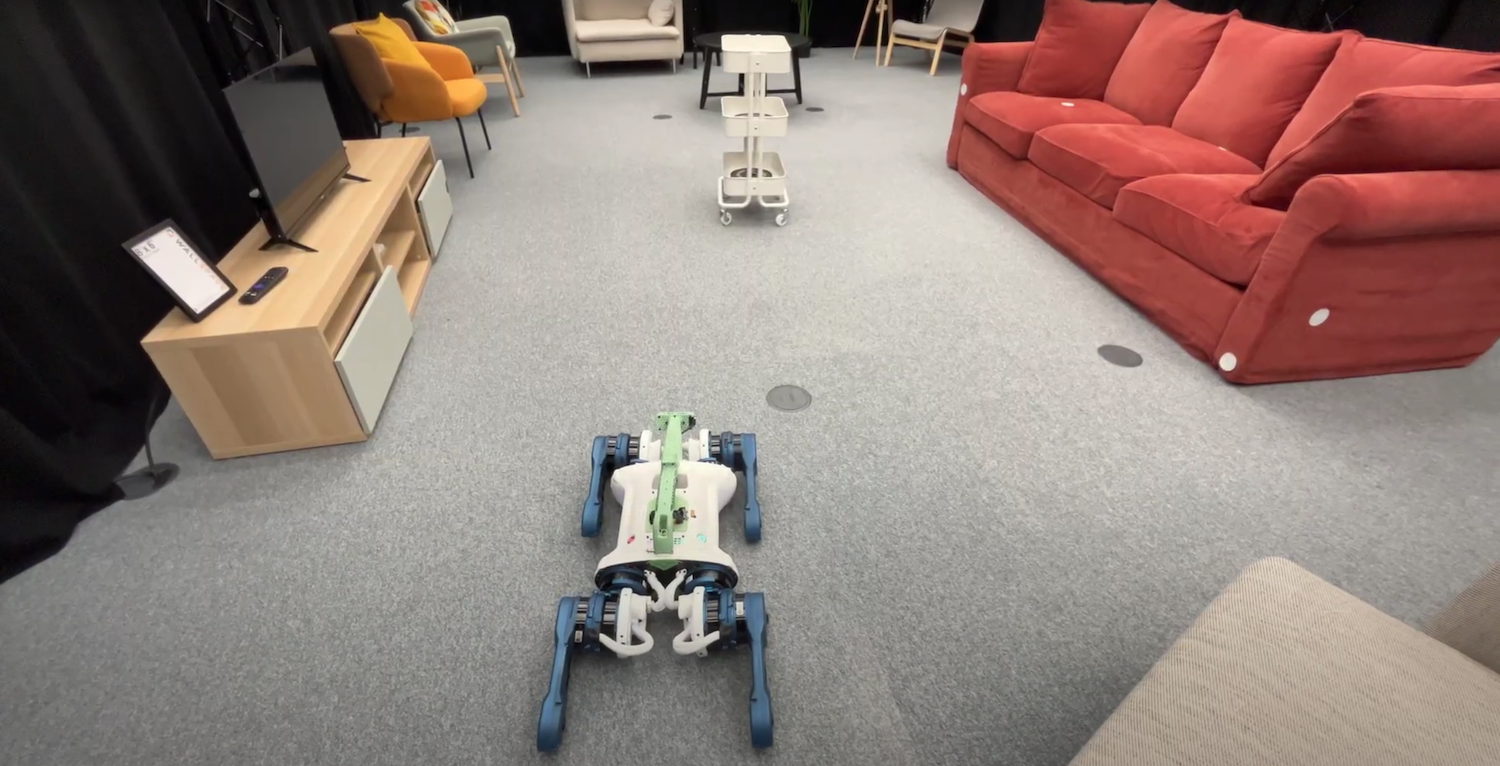}
        \caption{Trolley loaded with 10 kilograms of weights.}
    \end{subfigure}
    \caption{Out-of-distribution tests.}
    \label{fig:ood-targets}
\end{figure*}

\textbf{Do the policies recover the expert's behavior in simulation?}
Both the SPOC baseline and Gemini exhibit similarly strong overall performance and generalization in simulation, in the fixed-scene (Table \ref{tab:fixed-scene-results}) and procedurally-generated scenes (Figure \ref{fig:results-sim-procgen-scene}).
In particular, both policies recover much of privileged expert performance in the procedural generation scene, and generalize to both out-of-distribution text descriptions (with a small performance drop) and to a different language (with effectively no performance drop) despite being only trained on English descriptions.
This multilingual capability underscores the effectiveness of the highly diverse pretraining of both SigLIP \citep{zhai2023sigmoid}, the image/text encoder for the baseline trained on the Webli dataset  \citep{chen2022pali} containing 109 languages, and Gemini's strong multilingual capabilities.
The fixed scene was designed by humans, not created by a procedural generation recipe, so it is less cluttered and likely more representative of policy performance in the wild; this results in higher success rates relative to the procedural generation scene.

\textbf{How well do the policies transfer to hardware?}
In contrast to simulation, Gemini significantly outperforms the baseline on the more difficult hardware settings (Fig.~\ref{fig:results-hardware-with-initialization}). In particular, we see the baseline drops an average of 40\% in success rate in the hard settings compared to Gemini.
Furthermore, with the highly out-of-distribution target (giraffe plushie), the baseline has a 0\% success rate while Gemini achieves 70\%, indicating strong real-world out-of-distribution performance and robustness.
We also show Gemini has some ability to push the trolley to a human, a blue-and-white robot dog (i.e. a copy of its own hardware), and generalizes to the highly out-of-distribution physics setting where the trolley is loaded with 10 kilograms (22 pounds) of weights (see Fig.~\ref{fig:ood-targets} for visualizations of these settings).
This implies that the strength of Gemini pretraining combined with physics domain randomization results in robust behavior in the face of large distribution shifts.

\begin{figure}
    \centering
    \begin{subfigure}[b]{0.5\textwidth}
        \centering
        \includegraphics[width=0.8\textwidth]{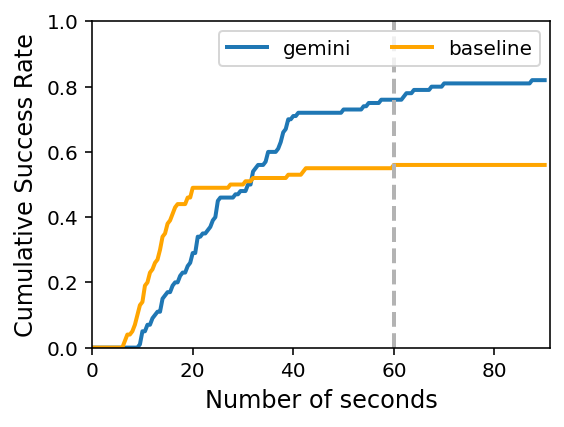}
    \end{subfigure}
    \caption{Cumulative success rate by episode time per policy. Only successes within 60 seconds count; this point is indicated by the dashed line.}
    \label{fig:cumulative_episode_length}
\end{figure}

Qualitatively, Gemini seems more cautious compared to the baseline.
See our website for videos of the policy acting in the real world: \href{https://sites.google.com/view/proc4gem}{sites.google.com/view/proc4gem}.
We provide some quantitative evidence of this behavior in Fig.~\ref{fig:cumulative_episode_length}, which shows that while baseline success rates are higher for episodes with lower number of steps, the Gemini policy is able to recover better and maintains robustness to the real-world scene, resulting in a significantly higher final success rate.

The required context length is highly task-dependent; somewhat counterintuitively, we observe that for our task, increasing the context beyond 8 time steps seems unnecessary. For example the baseline SPOC policy, trained with a doubled context of 16, achieves a comparable 55\% success on the train descriptions in procedurally-generated scenes.
However, a longer context is likely required for tasks that require navigation capability over many rooms, as described in \cite{ehsani2023imitating}; our chosen task stresses only short-to-medium distance navigation, but couples it with whole-body control.
Effective usage of the full context length of modern multimodal models for robotics is a fruitful area of future research.

%% file: related_work.tex
\section{Related work}

\begin{figure}
    \centering
    \includegraphics[width=0.7\textwidth]{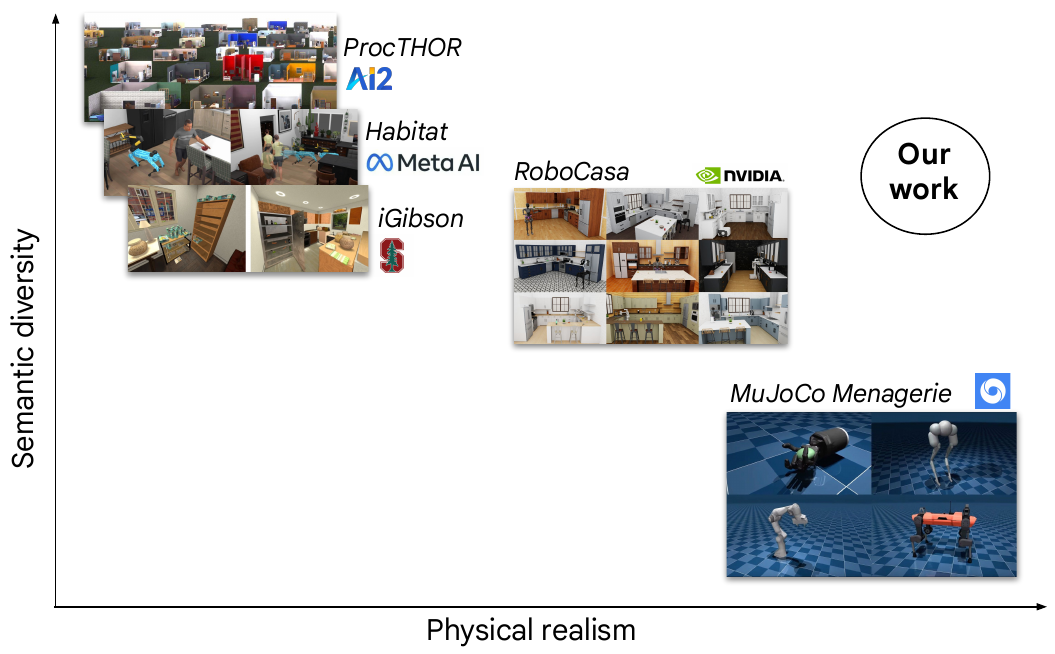}
    \caption{Comparison of simulations based on characteristics of typically-targeted tasks.}
    \label{fig:sims_comparison}
\end{figure}

\subsection{Sim2real in Robotics}
Sim2real transfer has made it possible to use reinforcement learning to solve several challenging real-world control problems~\citep{muratore2022robot,dimitropoulos2022brief}.
Careful system identification and techniques such as domain randomization~\citep{tobin2017domain}, domain adaptation~\citep{bousmalis2017unsupervised} and real2sim~\citep{chebotar2019closing} have helped to reduce the discrepancies between simulation and reality for the system dynamics and sensor model, enabling successes on tasks such as Rubik's cube solving with a dexterous hand~\citep{akkaya2019solving}, grasping~\citep{bousmalis2018using}, stacking~\citep{lee2021beyond}, autonomous flight~\citep{sadeghi2016cad2rl}, quadruped~\citep{hwangbo2019learning, Lee_2020, peng2020learning, sim2real-robotnpmp}, biped locomotion~\citep{yu2019sim, li2021reinforcement, siekmann2021blind}, biped navigation~\citep{byravan2023nerf2real} and robot soccer~\cite{haarnoja2024learning}. We rely on many of the lessons learned in these works, and propose a system for procedurally generating diverse indoor scenes with high quality physics and photo-realistic rendering, and show that this system can be used for sim2real transfer of whole body language conditioned object interactions.

\subsection{Expert distillation for sim2real}

Many previous works have used privileged expert distillation to turn policies trained on ground-truth simulation state to ones which only observe sensory data available to the robot. This has been especially successful for sim2real of blind \citep{lee2020challenging, kumar2021rma}, and depth or point cloud conditioned \citep{margolis2022jump, miki2022perceptive, agarwal2023legged, ziwen2023parkour, xuxin2024extreme} locomotion policies, depth-conditioned dexterous manipulation \citep{chen2022inhand, chen2024visual}, drone flight (using depth camera \cite{song2023cluttered} or bounding boxes from a pre-trained object detector \citep{song2023racing}), and simulated, autonomous driving \citep{chen2019learningcheating}.

Our work differs in several important aspects: 1. We use RGB instead of a depth camera, relying on high-fidelity, photorealistic simulation, and 2. We do not assume availability of waypoints or goal point clouds. Instead, our students are language-conditioned.

The closest to our work is \cite{ehsani2024spoc} which distills simulated trajectories from a motion-planner into a transformer-based vision and language conditioned student. Unlike this paper, we address a whole-body control task which not only requires mastering navigation but also reasoning about how and where to apply forces. We use the architecture from \cite{ehsani2024spoc} as our baseline.
Follow-up works \citep{zeng2024poliformer,hu2024flare} scale on-policy RL directly to the final policy, while we retain the BC approach but clone from RL-trained experts instead of planners.

\subsection{Embodied AI simulations and procedural generation}
Several simulation suites, such as Habitat~\citep{savva2019habitat, szot2021habitat, puig2023habitat}, (i)Gibson~\citep{xia2018gibson,xia2019gibson,shen2021igibson}, SAPIEN~\citep{xiang2020sapien}, AI2/ROBO-THOR~\citep{kolve2017ai2, deitke2020robothor}, ProcThor~\citep{deitke2022} and ThreeDWorld~\citep{gan2020threedworld}, have been proposed to tackle embodied AI tasks, combining 3D simulators with diverse databases of assets, scenes and embodiments~\citep{duan2022survey}. These simulators have been used for learning visual navigation policies~\citep{anderson2018evaluation, wijmans2019dd, chen2019learning}, solve object-based navigation~\citep{khandelwal2022simple, batra2020objectnav, gervet2023navigating} and to create benchmarks for navigation \& pick-and-place tasks in diverse environments~\citep{yenamandra2023homerobot, mu2021maniskill, gu2023maniskill2, li2023behavior, ehsani2024spoc}. Several of these works have shown interesting sim2real transfer leveraging realistic rendering pipelines, predominantly for visual navigation~\citep{gervet2023navigating} and combined navigation and pick-and-place tasks~\citep{yenamandra2023homerobot, ehsani2024spoc, gu2023maniskill2, li2023behavior}.

A key caveat in all these works is that the transfer is restricted often to non-interactive, or, at best, kinematic manipulation tasks leveraging simplified actions, or motion planning based primitives. In contrast, we show that recipes similar to that seen in prior work~\citep{li2023behavior, deitke2022} can be scaled all the way for sim2real transfer of highly interactive whole body control behaviors by leveraging procedural generation, high-quality physics~\citep{todorov2012mujoco} and photorealistic rendering~\citep{unity}.
While recent work also combines similar ingredients together for robot manipulation tasks~\citep{robocasa2024,gensim2024}, limited sim2real transfer has been shown from these simulation frameworks.

More recently, large language models (LLMs) have been used to create diverse procedural recipes for scene creation~\citep{yang2024holodeck}, as well as for task, scene and behavior generation~\citep{wang2023robogen, wang2023gensim}. Data augmentation approaches using diffusion models have also been used to increase scene diversity and realism, which can aid in better sim2real transfer~\citep{chen2023genaug, di2024diffusion, yu2023scaling}. These approaches can be used in tandem with our proposed method to further enhance the diversity of generated scenes and behaviors. Alternatively, state of the art vision models have also been used to create simulations from real-world scenes (real2sim)~\citep{byravan2023nerf2real, deitke2023phone2proc, shafiullah2022clip}. These are often created directly from the target scene where the robot is to be deployed, and hence do not generalise to novel scenes and are restricted to simple tasks such as navigation and pick-and-place. 

\subsection{Foundation models for robotic actions}

There has been significant recent progress in adding robotic action capabilities to foundation models, either by training them to directly output actions \citep{zeng2022socratic,reed2022generalist,lu2024unified,brohan2022rt,brohan2023rt,bousmalis2023robocat,team2024octo} or through an intermediate medium like language \citep{ahn2022can,driess2023palm} or code \citep{yu2023language,liang2023code,ma2023eureka}.
While primarily focused in the manipulation setting, \cite{shah2023lm,shah2023gnm,shah2023vint,chiang2024mobility} investigate foundation models for navigation, and the promise of cross-embodiment positive transfer inspires work which tackles both \citep{majumdar2023we,yang2024pushing,doshi2024scaling}.
Concurrent work on cross-embodiment, dexterous manipulation, and embodied reasoning through Gemini demonstrates complementary, exciting directions utilizing foundation models for robotics \citep{gemini_for_robotics}.
Limited by the availability of real-world data and the promise of cross embodiment transfer learning, scaling this line of work up has been driven by sharing data across institutions \citep{padalkar2023open,kim2024openvla}.
Our work hopes to complement this investment by the community by highlighting simulation as a source of highly diverse yet physically realistic data and its utility for training robot foundation models.

%% file: appendix.tex
\part*{Appendix}

\section{Simulation setup}
\paragraph{Asset placement}

\begin{figure*}
    \centering
    \includegraphics[width=0.95\textwidth]{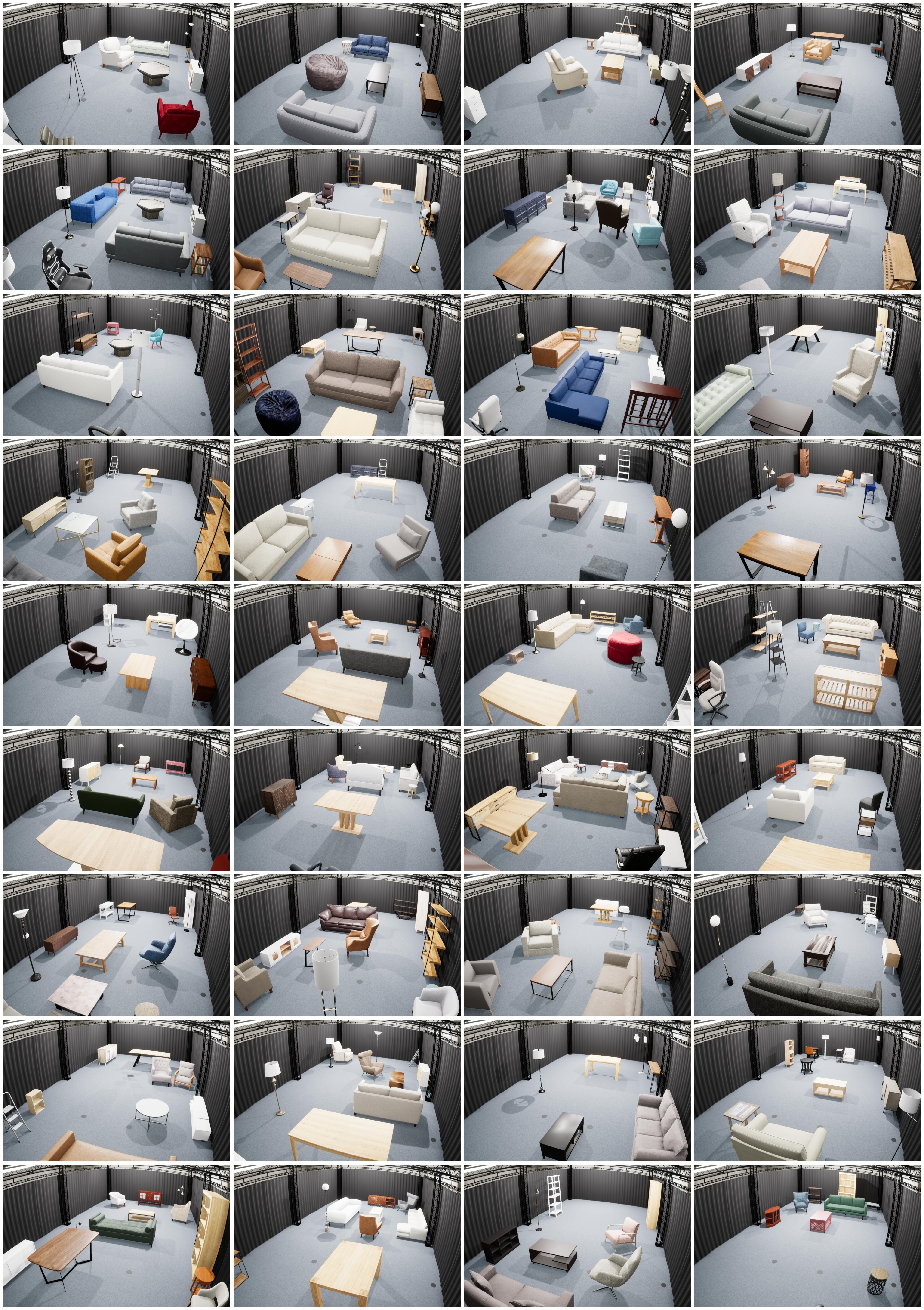}
    \caption{Example asset placements to create living room scenes.}
    \label{fig:living_room_samples_extended}
\end{figure*}

See Fig.~\ref{fig:living_room_samples_extended} for additional examples of rendered asset placements.
\label{sec:asset-placement-recipe}

\paragraph{Asset descriptions}

\begin{figure*}
    \centering
    \includegraphics[width=0.7\textwidth]{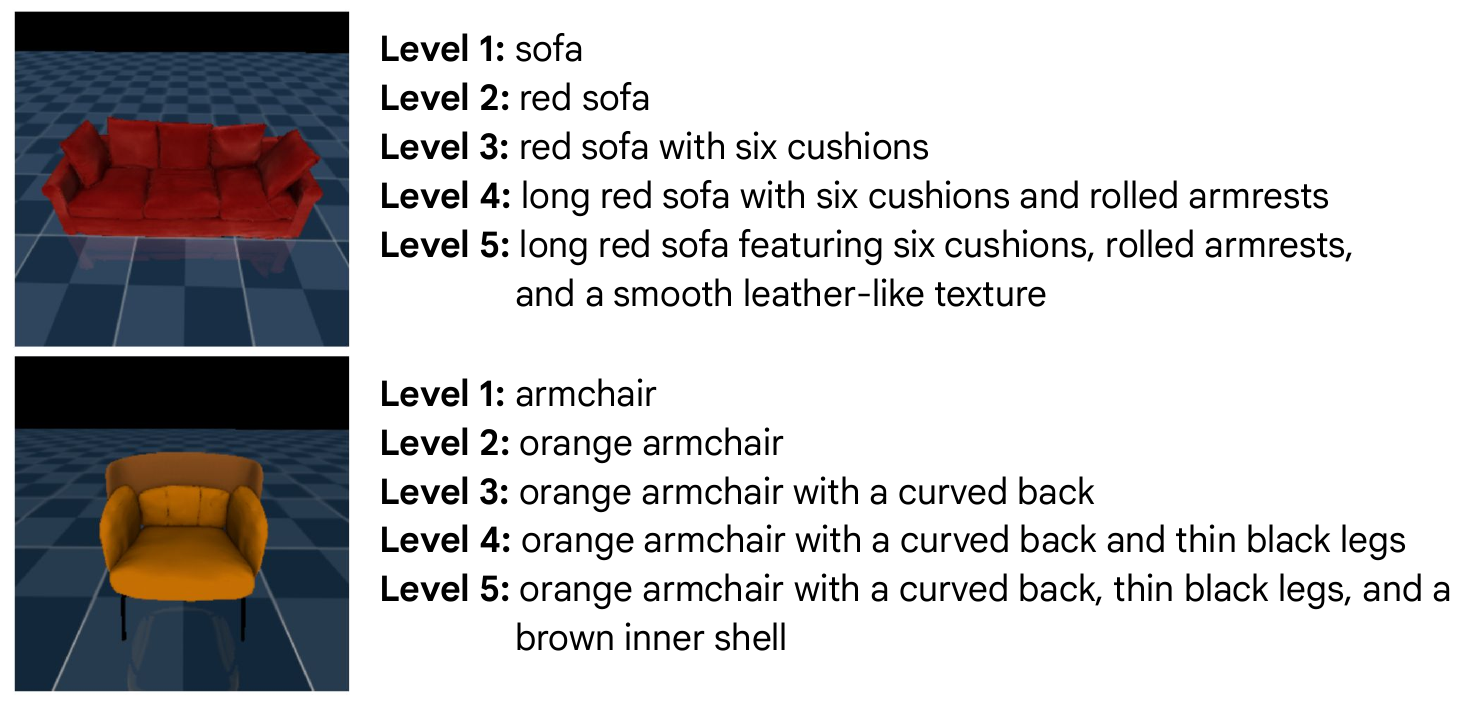}
    \caption{Example asset descriptions, for each of the five levels of complexity.}
    \label{fig:description_examples}
\end{figure*}

See Fig.~\ref{fig:description_examples} for examples of descriptions at each of the five complexity levels, generated by Gemini.

\section{Robot embodiment}

The Barkour robot \citep{caluwaerts2023barkour} is a small open source quadruped similar in size and weight to the Unitree A1 and MIT Mini-Cheetah quadrupeds, and has 3 degrees of freedom per leg and its standing height is about 40cm. 
We run the low-level locomotion controller on the robot's onboard computer (NUC11TNBv7) and connect to an off-robot workstation via WiFi.
The sensor suite for our experiments consist of joint encoders, an IMU (Microstrain 3DM-CV7-AHRS), and 2 onboard cameras: An Intel RealSense D435i located in the robot's head, and a Luxonis OAK-FFC-OV9782-M12 camera with an Arducam M23272M14 140 wide angle lens located on the robot's back.
See Figure~\ref{fig:barkour_robot} for a visualization of the robot.

\section{Hardware deployment}\label{appendix:hardware-deployment}

We evaluate the proposed agent on the real Barkour quadrupedal platform introduced in Sect.~\ref{section:platform}. The real-hardware evaluation setup consists of a distributed hierarchical control system setup, split into a lower level node and a higher level node. The low-level node is deployed locally on the robot PC and behaves as a closed loop system with the low-level locomotion controller described in Section~\ref{section:platform} running at $50\text{Hz}$. The high-level node runs on a workstation connected via wireless network to the robot. Similarly to the low-level node, also this nodes behaves as a closed loop system. The agent runs at $2\text{Hz}$ and communicates to a remote Gemini instance. 

\section{Results}
\begin{figure*}
    \begin{subfigure}[b]{\textwidth}
        \centering
        \includegraphics[width=\textwidth]{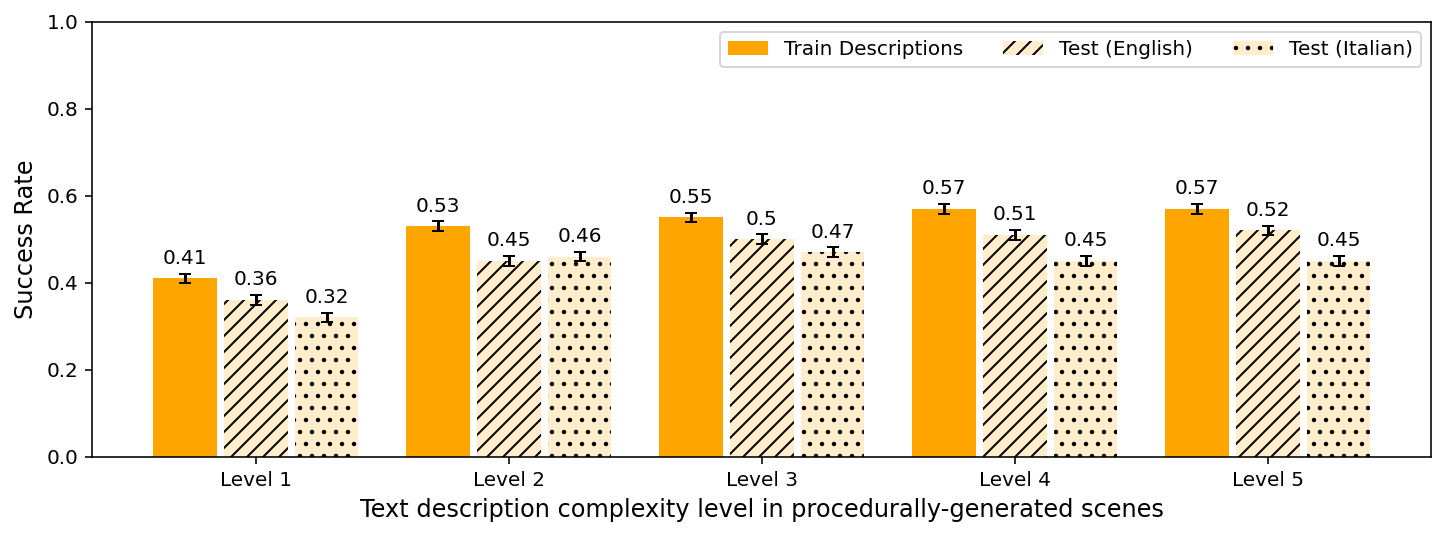}
        \caption{Gemini}
    \end{subfigure}
    \begin{subfigure}[b]{\textwidth}
        \centering
        \includegraphics[width=\textwidth]{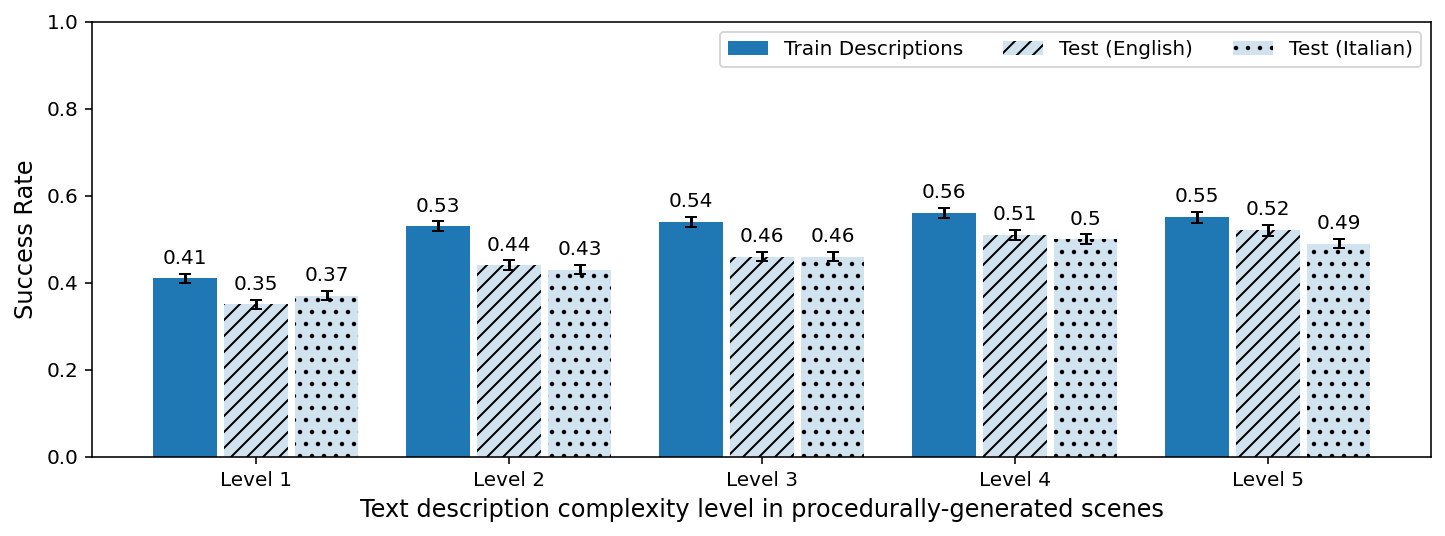}
        \caption{Baseline}
    \end{subfigure}
    \caption{Procedurally-generated simulation results, per description level. The error bars indicate standard error. The RL expert achieves 68.9\% success rate.}
    \label{fig:results-sim-procgen-scene-per-level}
\end{figure*}

See Figure~\ref{fig:results-sim-procgen-scene-per-level} for success rates for different levels of target object description detail, from simplest (level 1) to most detailed (level 5).

See Table~\ref{tab:hardware-eval-positions} for exact hardware initial conditions per target and difficulty level.

\begin{table*}
    \centering
    \begin{tabular}{|l|c c c|}
        \hline
        & \makecell{Easy \\ (trolley and target \\ in view throughout \\ episode)} & \makecell{Medium \\ (trolley and target \\ in view during \\ approach; target goes \\ out of view as robot \\ approaches trolley)} & \makecell{Hard \\ (trolley in view during \\ approach; target unseen)} \\
        \hline
        Red sofa & \includegraphics[width=0.22\textwidth]{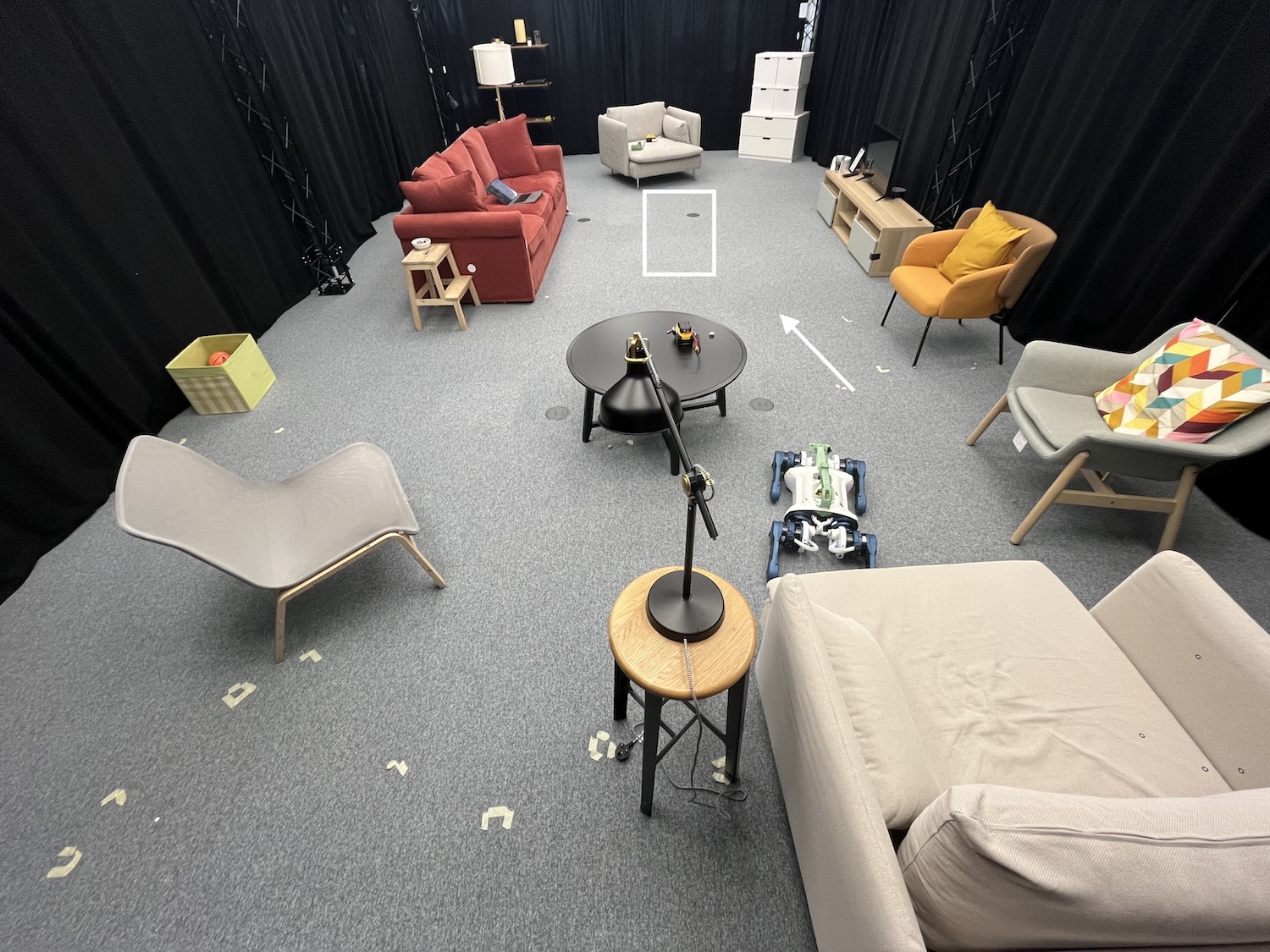} & \includegraphics[width=0.22\textwidth]{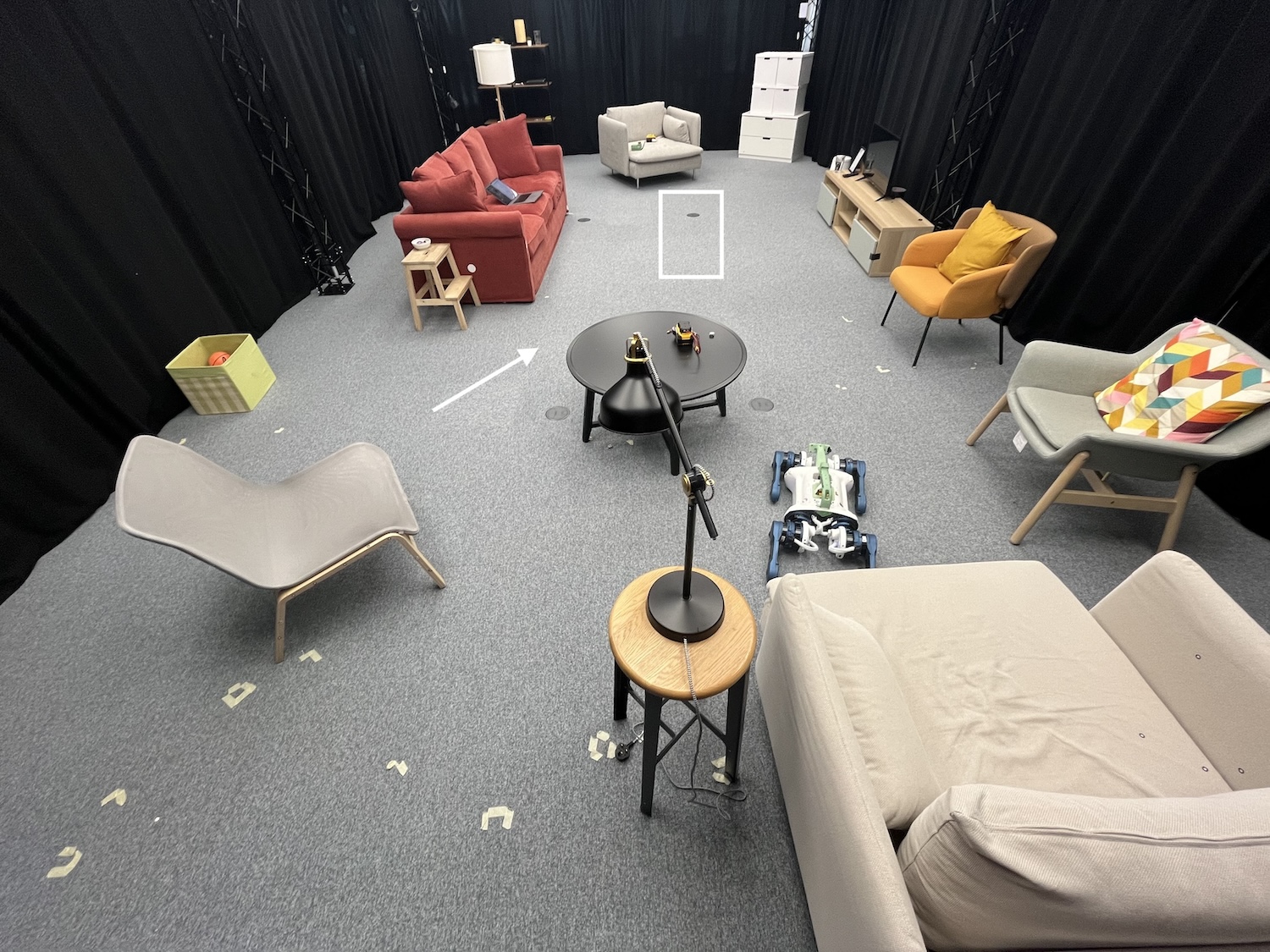} & \includegraphics[width=0.22\textwidth]{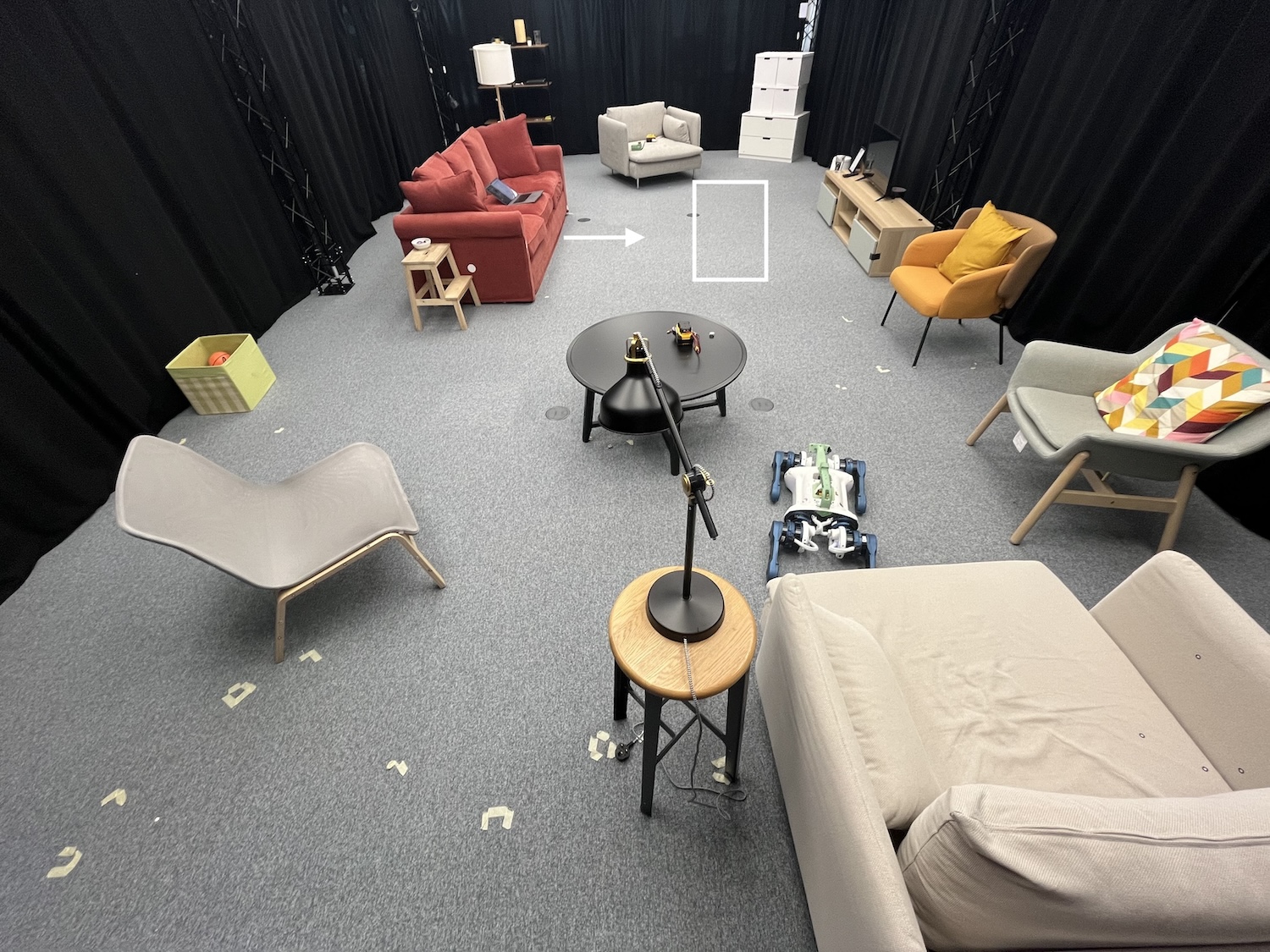}\\
        Orange armchair & \includegraphics[width=0.22\textwidth]{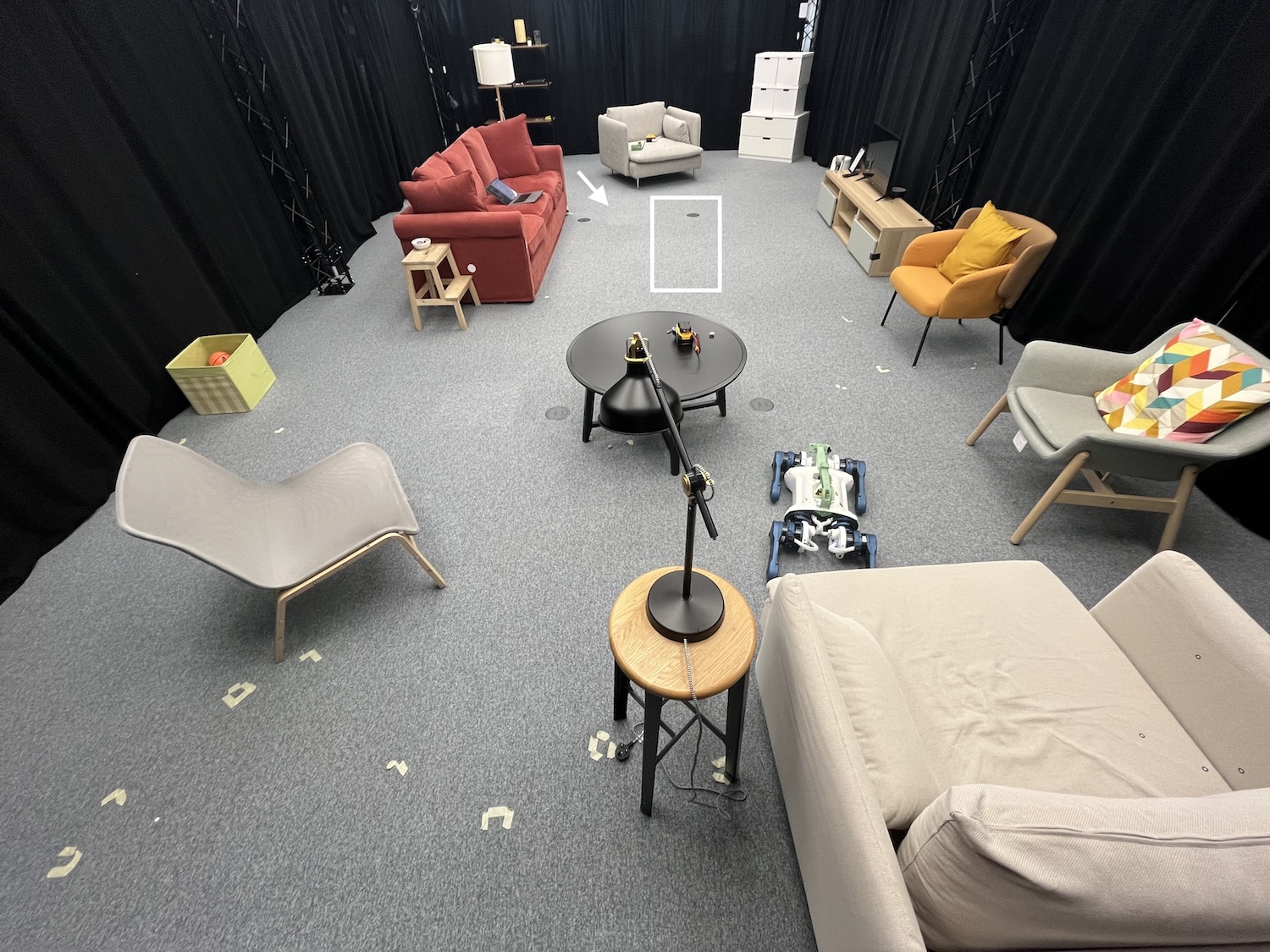} &  \includegraphics[width=0.22\textwidth]{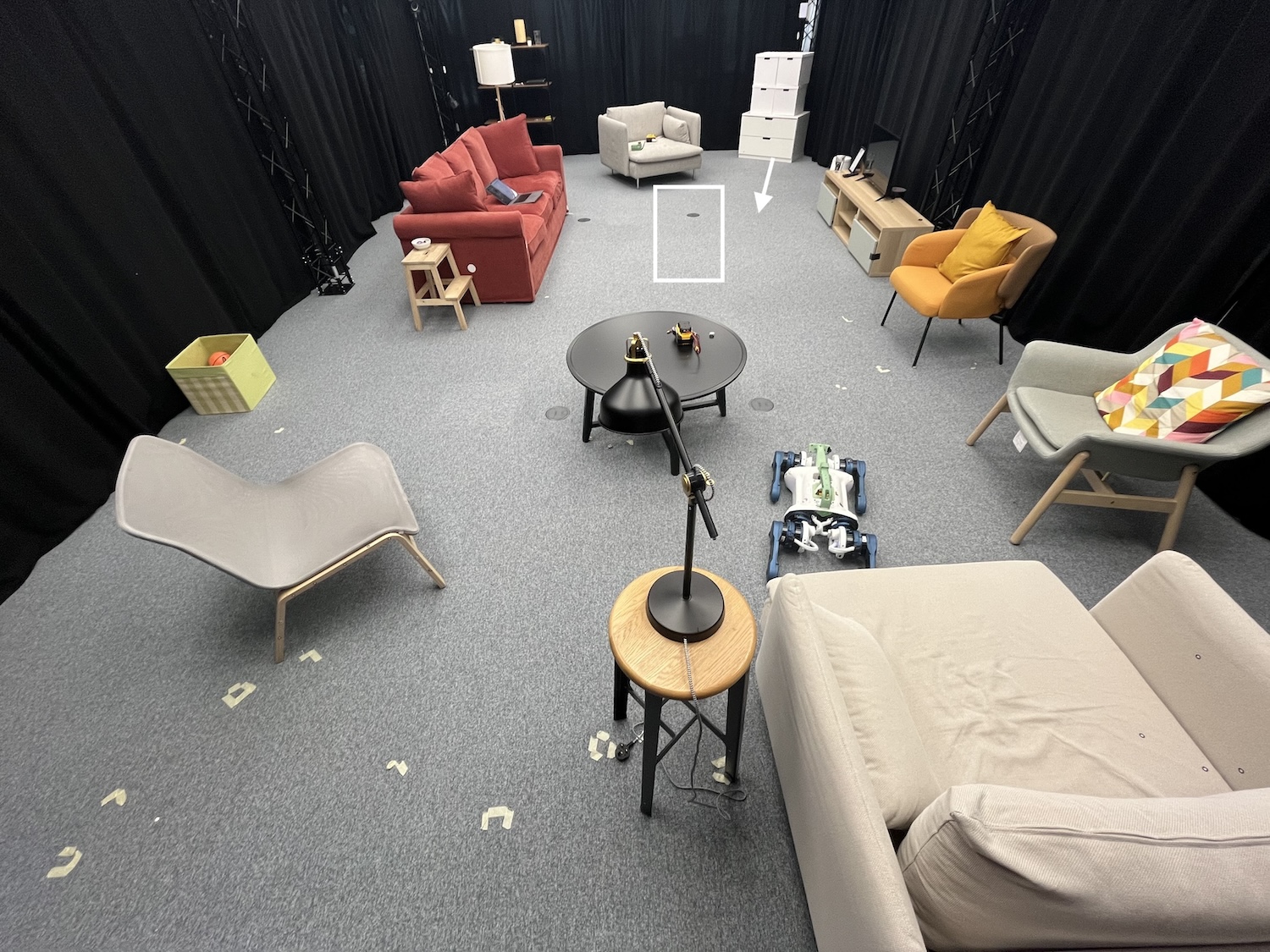} & \includegraphics[width=0.22\textwidth]{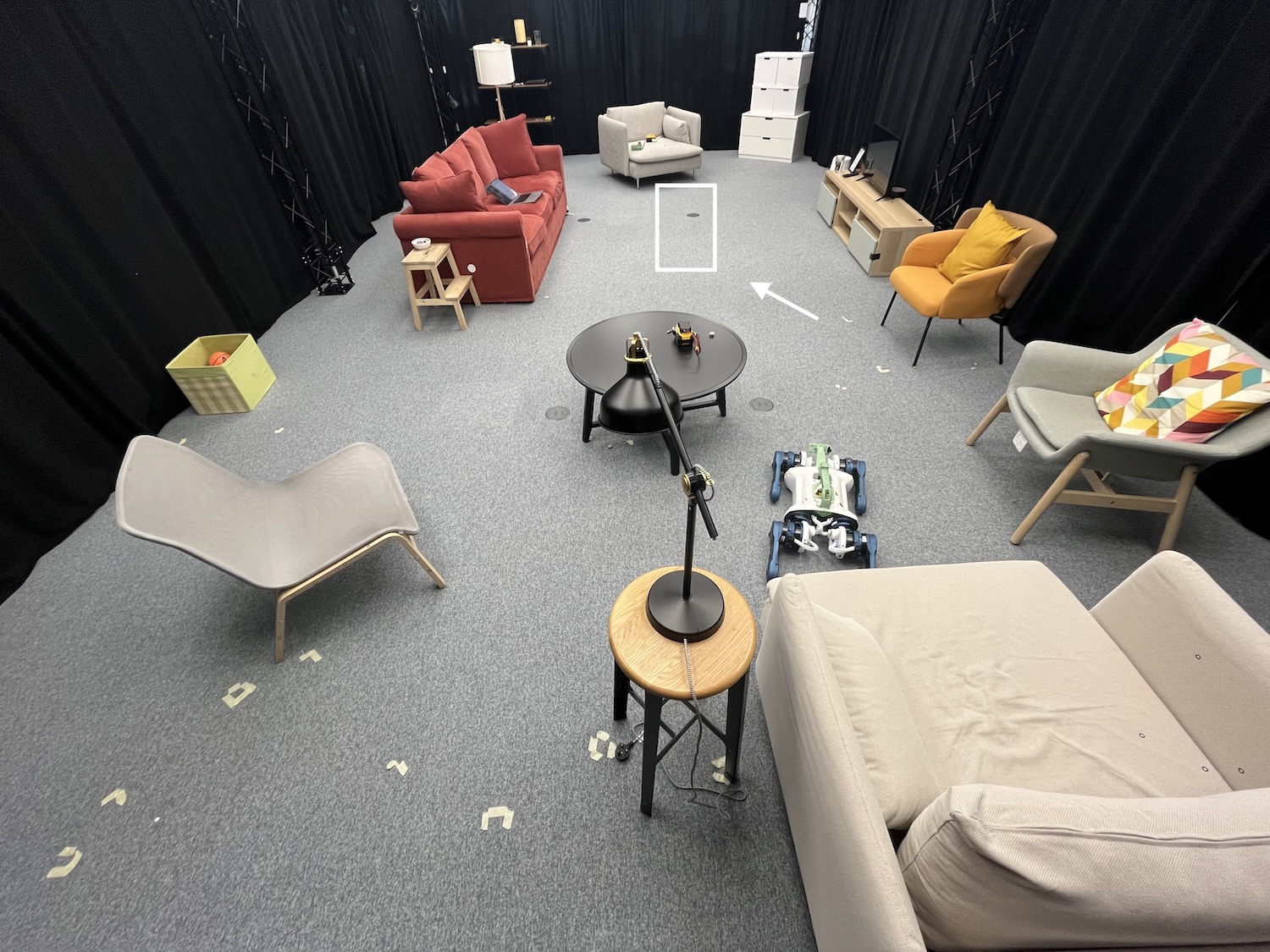}\\
        White bin & \includegraphics[width=0.22\textwidth]{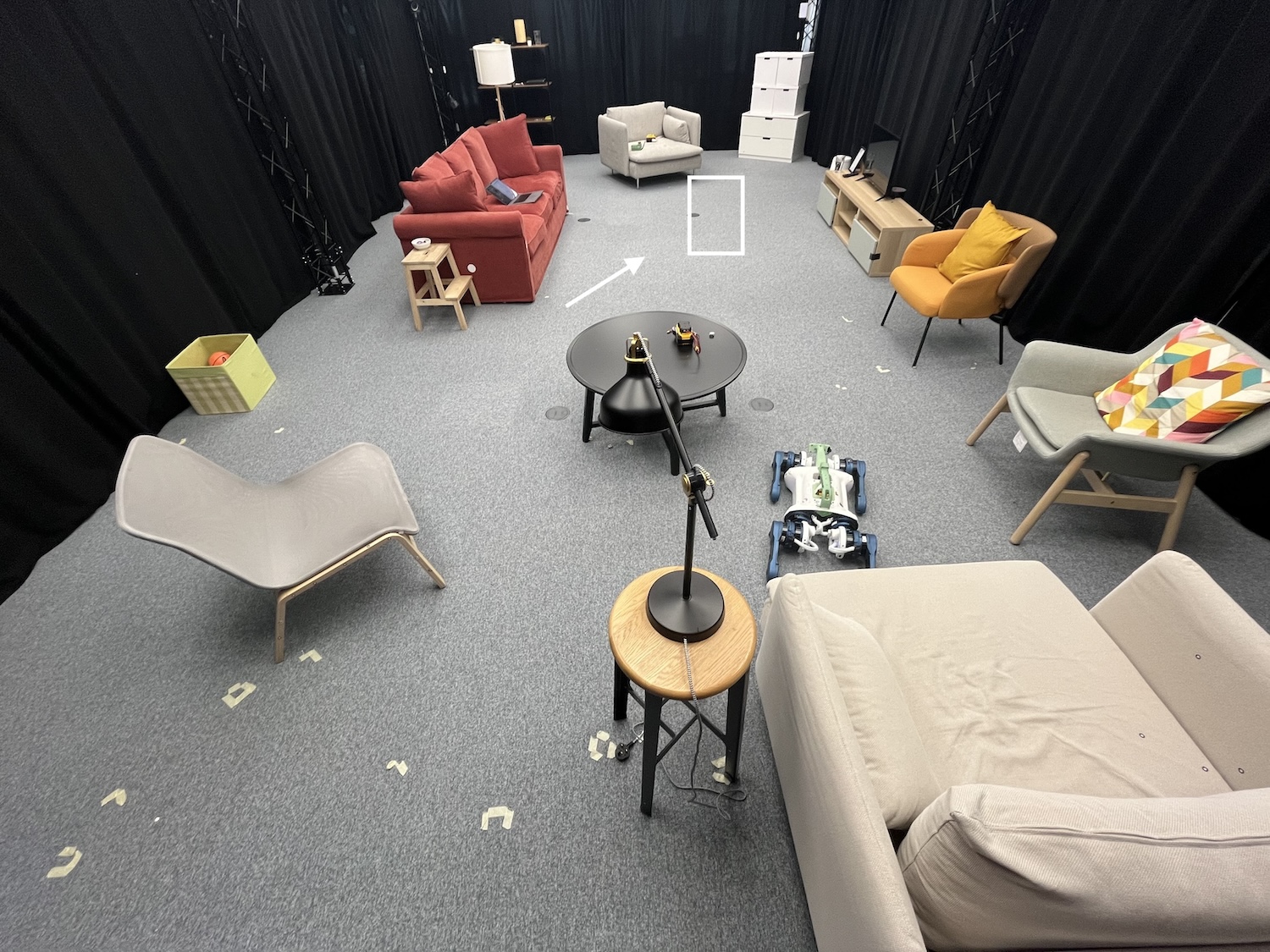} & \includegraphics[width=0.22\textwidth]{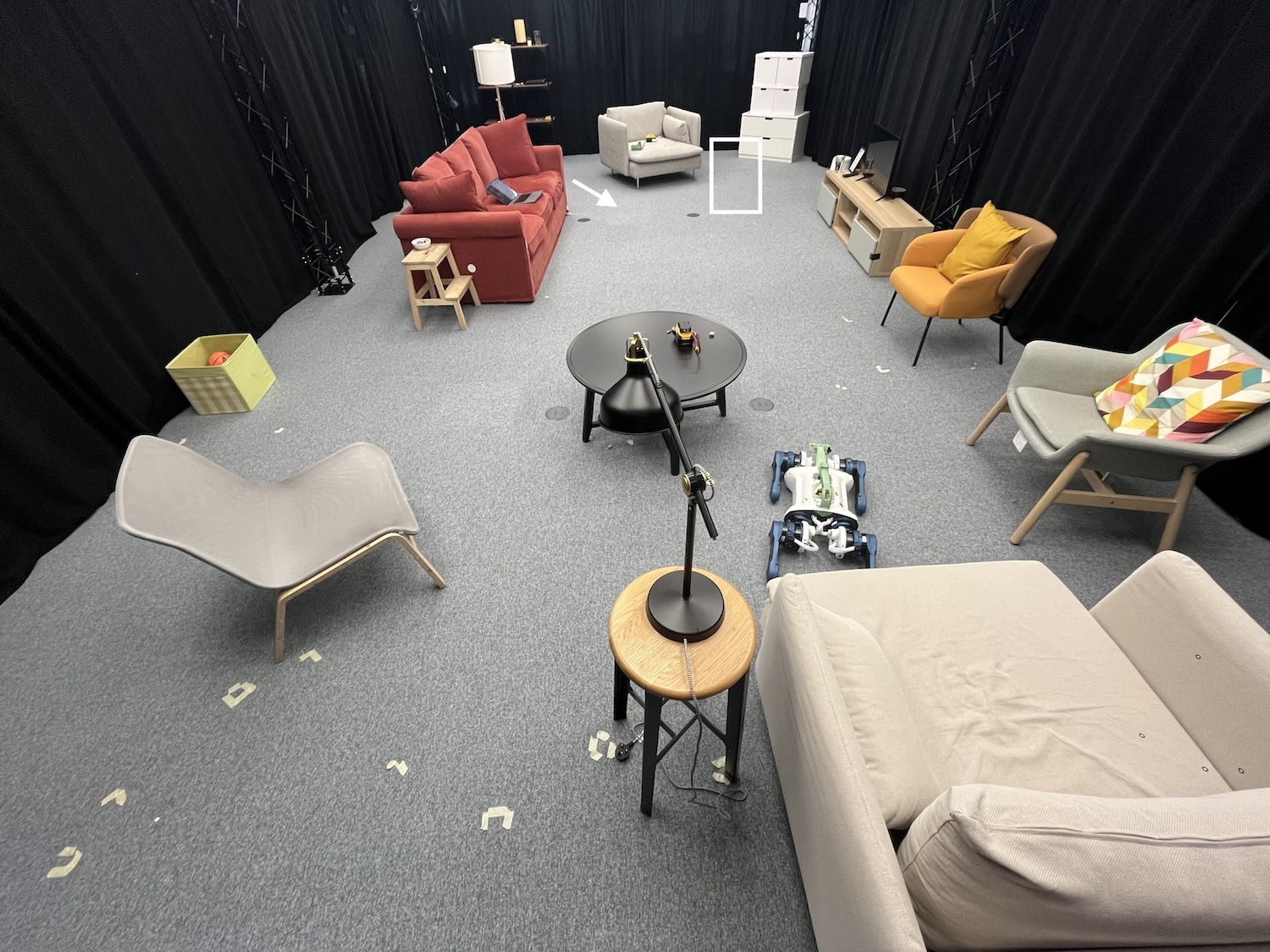} & \includegraphics[width=0.22\textwidth]{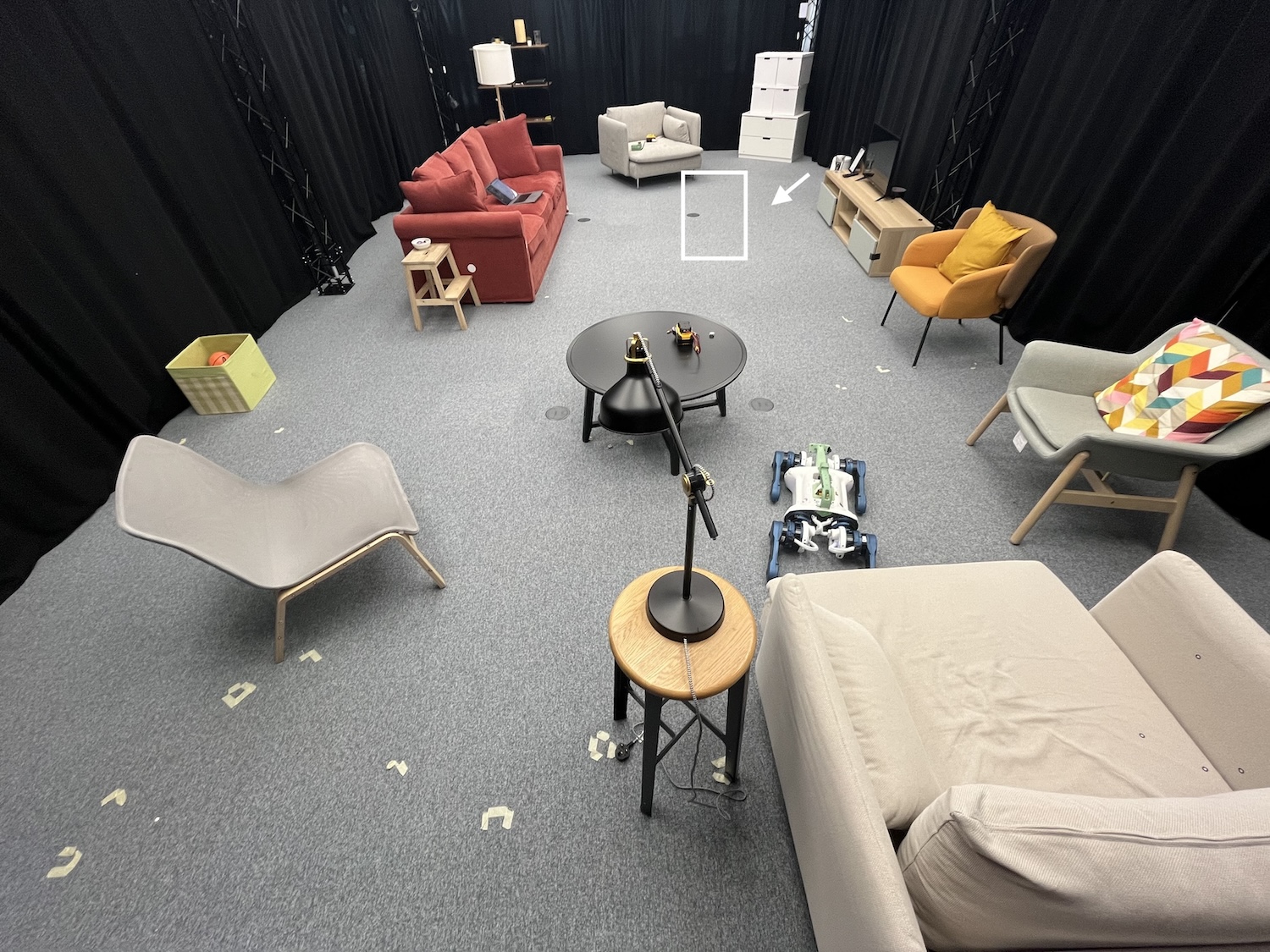} \\
        \hline
        \makecell{Giraffe \\ (exact positions \\ as in photo)} & & \includegraphics[width=0.22\textwidth]{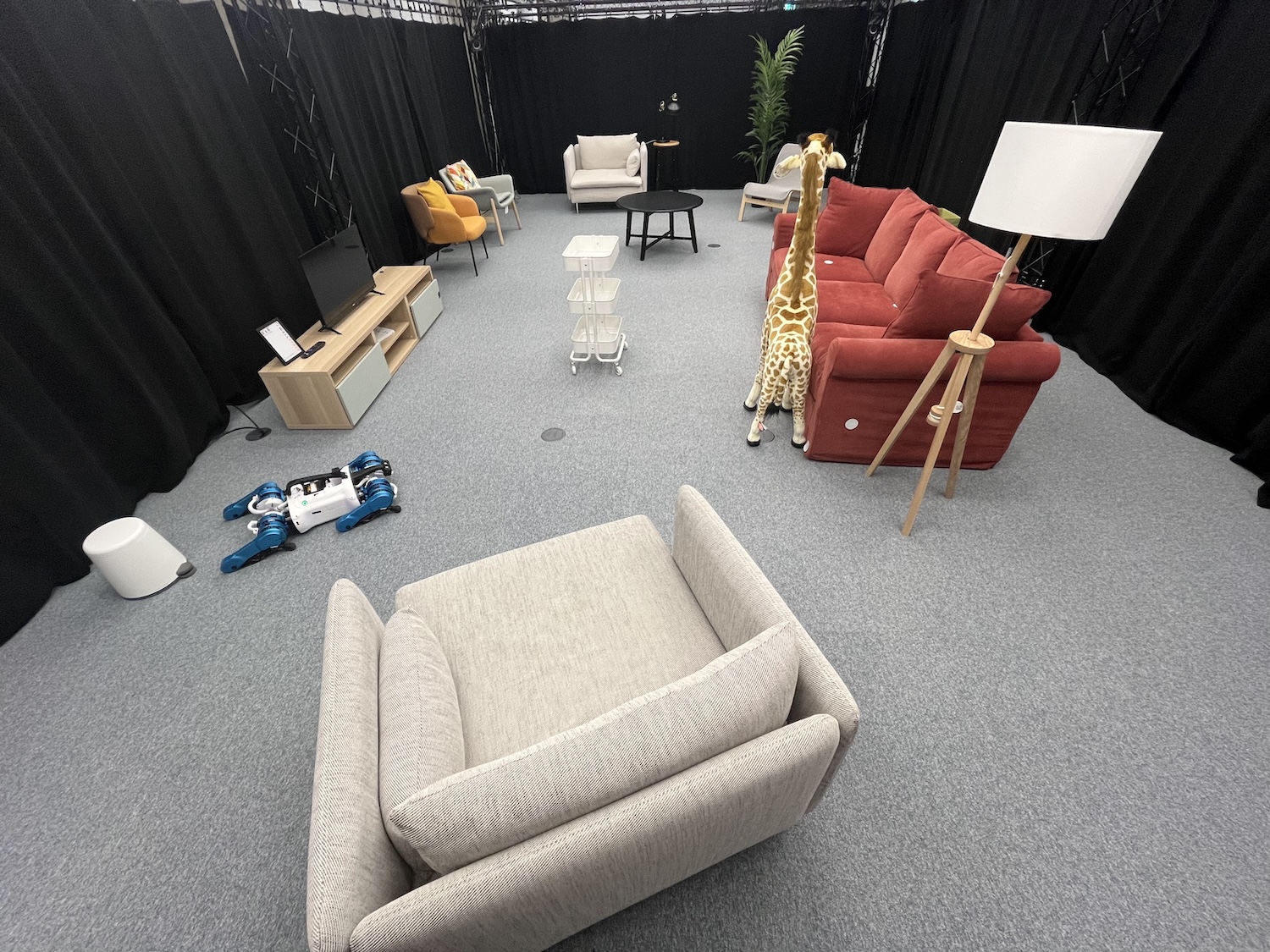} & \\
        \hline
    \end{tabular}
    \caption{Hardware evaluation initial conditions (positions and orientations for robot and trolley). White rectangle indicates trolley placement; white arrow indicates robot position and orientation.
    We define initial conditions to be \textbf{Easy} for when the robot can easily maintain both the trolley and target object within its field of view for the full episode, \textbf{Medium} for when the trolley and target object is initially in the field of view but the target leaves the field of view as the robot approaches the trolley, and finally \textbf{Hard} for the case in which the robot is initialized between the trolley and target and faces the trolley. These definitions map neatly to the final empirical success rates of our policies on hardware (see Fig.~\ref{fig:results-hardware-with-initialization})
    }
    \label{tab:hardware-eval-positions}
\end{table*}

%% file: 0_main_gdm.bbl
\begin{thebibliography}{110}
\providecommand{\natexlab}[1]{#1}
\providecommand{\url}[1]{\texttt{#1}}
\expandafter\ifx\csname urlstyle\endcsname\relax
  \providecommand{\doi}[1]{doi: #1}\else
  \providecommand{\doi}{doi: \begingroup \urlstyle{rm}\Url}\fi

\bibitem[Achiam et~al.(2023)Achiam, Adler, Agarwal, Ahmad, Akkaya, Aleman,
  Almeida, Altenschmidt, Altman, Anadkat, et~al.]{achiam2023gpt}
J.~Achiam, S.~Adler, S.~Agarwal, L.~Ahmad, I.~Akkaya, F.~L. Aleman, D.~Almeida,
  J.~Altenschmidt, S.~Altman, S.~Anadkat, et~al.
\newblock Gpt-4 technical report.
\newblock \emph{arXiv preprint arXiv:2303.08774}, 2023.

\bibitem[Agarwal et~al.(2023)Agarwal, Kumar, Malik, and
  Pathak]{agarwal2023legged}
A.~Agarwal, A.~Kumar, J.~Malik, and D.~Pathak.
\newblock Legged locomotion in challenging terrains using egocentric vision.
\newblock In \emph{Conference on robot learning}, pages 403--415. PMLR, 2023.

\bibitem[Ahn et~al.(2022)Ahn, Brohan, Brown, Chebotar, Cortes, David, Finn, Fu,
  Gopalakrishnan, Hausman, et~al.]{ahn2022can}
M.~Ahn, A.~Brohan, N.~Brown, Y.~Chebotar, O.~Cortes, B.~David, C.~Finn, C.~Fu,
  K.~Gopalakrishnan, K.~Hausman, et~al.
\newblock Do as i can, not as i say: Grounding language in robotic affordances.
\newblock \emph{arXiv preprint arXiv:2204.01691}, 2022.

\bibitem[Akkaya et~al.(2019)Akkaya, Andrychowicz, Chociej, Litwin, McGrew,
  Petron, Paino, Plappert, Powell, Ribas, et~al.]{akkaya2019solving}
I.~Akkaya, M.~Andrychowicz, M.~Chociej, M.~Litwin, B.~McGrew, A.~Petron,
  A.~Paino, M.~Plappert, G.~Powell, R.~Ribas, et~al.
\newblock Solving rubik's cube with a robot hand.
\newblock \emph{arXiv preprint arXiv:1910.07113}, 2019.

\bibitem[Anderson et~al.(2018)Anderson, Chang, Chaplot, Dosovitskiy, Gupta,
  Koltun, Kosecka, Malik, Mottaghi, Savva, et~al.]{anderson2018evaluation}
P.~Anderson, A.~Chang, D.~S. Chaplot, A.~Dosovitskiy, S.~Gupta, V.~Koltun,
  J.~Kosecka, J.~Malik, R.~Mottaghi, M.~Savva, et~al.
\newblock On evaluation of embodied navigation agents.
\newblock \emph{arXiv preprint arXiv:1807.06757}, 2018.

\bibitem[Batra et~al.(2020)Batra, Gokaslan, Kembhavi, Maksymets, Mottaghi,
  Savva, Toshev, and Wijmans]{batra2020objectnav}
D.~Batra, A.~Gokaslan, A.~Kembhavi, O.~Maksymets, R.~Mottaghi, M.~Savva,
  A.~Toshev, and E.~Wijmans.
\newblock Objectnav revisited: On evaluation of embodied agents navigating to
  objects.
\newblock \emph{arXiv preprint arXiv:2006.13171}, 2020.

\bibitem[Bauza et~al.(2024)Bauza, Chen, Dalibard, Gileadi, Hafner, Martins,
  Moore, Pevceviciute, Laurens, Rao, et~al.]{bauza2024demostart}
M.~Bauza, J.~E. Chen, V.~Dalibard, N.~Gileadi, R.~Hafner, M.~F. Martins,
  J.~Moore, R.~Pevceviciute, A.~Laurens, D.~Rao, et~al.
\newblock Demostart: Demonstration-led auto-curriculum applied to sim-to-real
  with multi-fingered robots.
\newblock \emph{arXiv preprint arXiv:2409.06613}, 2024.

\bibitem[Bohez et~al.(2022)Bohez, Tunyasuvunakool, Brakel, Sadeghi,
  Hasenclever, Tassa, Parisotto, Humplik, Haarnoja, Hafner, Wulfmeier, Neunert,
  Moran, Siegel, Huber, Romano, Batchelor, Casarini, Merel, Hadsell, and
  Heess]{sim2real-robotnpmp}
S.~Bohez, S.~Tunyasuvunakool, P.~Brakel, F.~Sadeghi, L.~Hasenclever, Y.~Tassa,
  E.~Parisotto, J.~Humplik, T.~Haarnoja, R.~Hafner, M.~Wulfmeier, M.~Neunert,
  B.~Moran, N.~Siegel, A.~Huber, F.~Romano, N.~Batchelor, F.~Casarini,
  J.~Merel, R.~Hadsell, and N.~Heess.
\newblock Imitate and repurpose: Learning reusable robot movement skills from
  human and animal behaviors.
\newblock \emph{arXiv}, Mar. 2022.

\bibitem[Bousmalis et~al.(2017)Bousmalis, Silberman, Dohan, Erhan, and
  Krishnan]{bousmalis2017unsupervised}
K.~Bousmalis, N.~Silberman, D.~Dohan, D.~Erhan, and D.~Krishnan.
\newblock Unsupervised pixel-level domain adaptation with generative
  adversarial networks.
\newblock In \emph{Proceedings of the IEEE conference on computer vision and
  pattern recognition}, pages 3722--3731, 2017.

\bibitem[Bousmalis et~al.(2018)Bousmalis, Irpan, Wohlhart, Bai, Kelcey,
  Kalakrishnan, Downs, Ibarz, Pastor, Konolige, et~al.]{bousmalis2018using}
K.~Bousmalis, A.~Irpan, P.~Wohlhart, Y.~Bai, M.~Kelcey, M.~Kalakrishnan,
  L.~Downs, J.~Ibarz, P.~Pastor, K.~Konolige, et~al.
\newblock Using simulation and domain adaptation to improve efficiency of deep
  robotic grasping.
\newblock In \emph{2018 IEEE international conference on robotics and
  automation (ICRA)}, pages 4243--4250. IEEE, 2018.

\bibitem[Bousmalis et~al.(2023)Bousmalis, Vezzani, Rao, Devin, Lee, Bauza,
  Davchev, Zhou, Gupta, Raju, et~al.]{bousmalis2023robocat}
K.~Bousmalis, G.~Vezzani, D.~Rao, C.~Devin, A.~X. Lee, M.~Bauza, T.~Davchev,
  Y.~Zhou, A.~Gupta, A.~Raju, et~al.
\newblock Robocat: A self-improving foundation agent for robotic manipulation.
\newblock \emph{arXiv preprint arXiv:2306.11706}, 2023.

\bibitem[Brohan et~al.(2022)Brohan, Brown, Carbajal, Chebotar, Dabis, Finn,
  Gopalakrishnan, Hausman, Herzog, Hsu, et~al.]{brohan2022rt}
A.~Brohan, N.~Brown, J.~Carbajal, Y.~Chebotar, J.~Dabis, C.~Finn,
  K.~Gopalakrishnan, K.~Hausman, A.~Herzog, J.~Hsu, et~al.
\newblock Rt-1: Robotics transformer for real-world control at scale.
\newblock \emph{arXiv preprint arXiv:2212.06817}, 2022.

\bibitem[Brohan et~al.(2023)Brohan, Brown, Carbajal, Chebotar, Chen,
  Choromanski, Ding, Driess, Dubey, Finn, et~al.]{brohan2023rt}
A.~Brohan, N.~Brown, J.~Carbajal, Y.~Chebotar, X.~Chen, K.~Choromanski,
  T.~Ding, D.~Driess, A.~Dubey, C.~Finn, et~al.
\newblock Rt-2: Vision-language-action models transfer web knowledge to robotic
  control.
\newblock \emph{arXiv preprint arXiv:2307.15818}, 2023.

\bibitem[Byravan et~al.(2023)Byravan, Humplik, Hasenclever, Brussee, Nori,
  Haarnoja, Moran, Bohez, Sadeghi, Vujatovic, et~al.]{byravan2023nerf2real}
A.~Byravan, J.~Humplik, L.~Hasenclever, A.~Brussee, F.~Nori, T.~Haarnoja,
  B.~Moran, S.~Bohez, F.~Sadeghi, B.~Vujatovic, et~al.
\newblock Nerf2real: Sim2real transfer of vision-guided bipedal motion skills
  using neural radiance fields.
\newblock In \emph{2023 IEEE International Conference on Robotics and
  Automation (ICRA)}, pages 9362--9369. IEEE, 2023.

\bibitem[Caluwaerts et~al.(2023)Caluwaerts, Iscen, Kew, Yu, Zhang, Freeman,
  Lee, Lee, Saliceti, Zhuang, et~al.]{caluwaerts2023barkour}
K.~Caluwaerts, A.~Iscen, J.~C. Kew, W.~Yu, T.~Zhang, D.~Freeman, K.-H. Lee,
  L.~Lee, S.~Saliceti, V.~Zhuang, et~al.
\newblock Barkour: Benchmarking animal-level agility with quadruped robots.
\newblock \emph{arXiv preprint arXiv:2305.14654}, 2023.

\bibitem[Chang et~al.(2017)Chang, Dai, Funkhouser, Halber, Niessner, Savva,
  Song, Zeng, and Zhang]{chang2017matterport3d}
A.~Chang, A.~Dai, T.~Funkhouser, M.~Halber, M.~Niessner, M.~Savva, S.~Song,
  A.~Zeng, and Y.~Zhang.
\newblock Matterport3d: Learning from rgb-d data in indoor environments.
\newblock \emph{arXiv preprint arXiv:1709.06158}, 2017.

\bibitem[Chebotar et~al.(2019)Chebotar, Handa, Makoviychuk, Macklin, Issac,
  Ratliff, and Fox]{chebotar2019closing}
Y.~Chebotar, A.~Handa, V.~Makoviychuk, M.~Macklin, J.~Issac, N.~Ratliff, and
  D.~Fox.
\newblock Closing the sim-to-real loop: Adapting simulation randomization with
  real world experience.
\newblock In \emph{2019 International Conference on Robotics and Automation
  (ICRA)}, pages 8973--8979. IEEE, 2019.

\bibitem[Chen et~al.(2019{\natexlab{a}})Chen, Zhou, Koltun, and
  Krähenbühl]{chen2019learningcheating}
D.~Chen, B.~Zhou, V.~Koltun, and P.~Krähenbühl.
\newblock Learning by cheating, 2019{\natexlab{a}}.
\newblock URL \url{https://arxiv.org/abs/1912.12294}.

\bibitem[Chen et~al.(2019{\natexlab{b}})Chen, Gupta, and
  Gupta]{chen2019learning}
T.~Chen, S.~Gupta, and A.~Gupta.
\newblock Learning exploration policies for navigation.
\newblock \emph{arXiv preprint arXiv:1903.01959}, 2019{\natexlab{b}}.

\bibitem[Chen et~al.(2022{\natexlab{a}})Chen, Xu, and Agrawal]{chen2022inhand}
T.~Chen, J.~Xu, and P.~Agrawal.
\newblock A system for general in-hand object re-orientation.
\newblock In A.~Faust, D.~Hsu, and G.~Neumann, editors, \emph{Proceedings of
  the 5th Conference on Robot Learning}, volume 164 of \emph{Proceedings of
  Machine Learning Research}, pages 297--307. PMLR, 08--11 Nov
  2022{\natexlab{a}}.
\newblock URL \url{https://proceedings.mlr.press/v164/chen22a.html}.

\bibitem[Chen et~al.(2023{\natexlab{a}})Chen, Tippur, Wu, Kumar, Adelson, and
  Agrawal]{chen2024visual}
T.~Chen, M.~Tippur, S.~Wu, V.~Kumar, E.~Adelson, and P.~Agrawal.
\newblock Visual dexterity: In-hand reorientation of novel and complex object
  shapes.
\newblock \emph{Science Robotics}, 8\penalty0 (84):\penalty0 eadc9244,
  2023{\natexlab{a}}.
\newblock \doi{10.1126/scirobotics.adc9244}.
\newblock URL
  \url{https://www.science.org/doi/abs/10.1126/scirobotics.adc9244}.

\bibitem[Chen et~al.(2022{\natexlab{b}})Chen, Wang, Changpinyo, Piergiovanni,
  Padlewski, Salz, Goodman, Grycner, Mustafa, Beyer, et~al.]{chen2022pali}
X.~Chen, X.~Wang, S.~Changpinyo, A.~Piergiovanni, P.~Padlewski, D.~Salz,
  S.~Goodman, A.~Grycner, B.~Mustafa, L.~Beyer, et~al.
\newblock Pali: A jointly-scaled multilingual language-image model.
\newblock \emph{arXiv preprint arXiv:2209.06794}, 2022{\natexlab{b}}.

\bibitem[Chen et~al.(2023{\natexlab{b}})Chen, Kiami, Gupta, and
  Kumar]{chen2023genaug}
Z.~Chen, S.~Kiami, A.~Gupta, and V.~Kumar.
\newblock Genaug: Retargeting behaviors to unseen situations via generative
  augmentation.
\newblock \emph{arXiv preprint arXiv:2302.06671}, 2023{\natexlab{b}}.

\bibitem[Cheng et~al.(2023)Cheng, Kumar, and Pathak]{cheng2023legs}
X.~Cheng, A.~Kumar, and D.~Pathak.
\newblock Legs as manipulator: Pushing quadrupedal agility beyond locomotion.
\newblock In \emph{2023 IEEE International Conference on Robotics and
  Automation (ICRA)}, pages 5106--5112. IEEE, 2023.

\bibitem[Cheng et~al.(2024)Cheng, Shi, Agarwal, and Pathak]{xuxin2024extreme}
X.~Cheng, K.~Shi, A.~Agarwal, and D.~Pathak.
\newblock Extreme parkour with legged robots.
\newblock In \emph{2024 IEEE International Conference on Robotics and
  Automation (ICRA)}, pages 11443--11450, 2024.
\newblock \doi{10.1109/ICRA57147.2024.10610200}.

\bibitem[Chiang et~al.(2024)Chiang, Xu, Fu, Jacob, Zhang, Lee, Yu, Schenck,
  Rendleman, Shah, et~al.]{chiang2024mobility}
H.-T.~L. Chiang, Z.~Xu, Z.~Fu, M.~G. Jacob, T.~Zhang, T.-W.~E. Lee, W.~Yu,
  C.~Schenck, D.~Rendleman, D.~Shah, et~al.
\newblock Mobility vla: Multimodal instruction navigation with long-context
  vlms and topological graphs.
\newblock \emph{arXiv preprint arXiv:2407.07775}, 2024.

\bibitem[Coumans and Bai(2016--2021)]{coumans2021}
E.~Coumans and Y.~Bai.
\newblock Pybullet, a python module for physics simulation for games, robotics
  and machine learning.
\newblock \url{http://pybullet.org}, 2016--2021.

\bibitem[Deitke et~al.(2020)Deitke, Han, Herrasti, Kembhavi, Kolve, Mottaghi,
  Salvador, Schwenk, VanderBilt, Wallingford, et~al.]{deitke2020robothor}
M.~Deitke, W.~Han, A.~Herrasti, A.~Kembhavi, E.~Kolve, R.~Mottaghi,
  J.~Salvador, D.~Schwenk, E.~VanderBilt, M.~Wallingford, et~al.
\newblock Robothor: An open simulation-to-real embodied ai platform.
\newblock In \emph{Proceedings of the IEEE/CVF conference on computer vision
  and pattern recognition}, pages 3164--3174, 2020.

\bibitem[Deitke et~al.(2022)Deitke, VanderBilt, Herrasti, Weihs, Ehsani,
  Salvador, Han, Kolve, Kembhavi, and Mottaghi]{deitke2022}
M.~Deitke, E.~VanderBilt, A.~Herrasti, L.~Weihs, K.~Ehsani, J.~Salvador,
  W.~Han, E.~Kolve, A.~Kembhavi, and R.~Mottaghi.
\newblock Procthor: Large-scale embodied ai using procedural generation.
\newblock \emph{Advances in Neural Information Processing Systems}, 2022.

\bibitem[Deitke et~al.(2023)Deitke, Hendrix, Farhadi, Ehsani, and
  Kembhavi]{deitke2023phone2proc}
M.~Deitke, R.~Hendrix, A.~Farhadi, K.~Ehsani, and A.~Kembhavi.
\newblock Phone2proc: Bringing robust robots into our chaotic world.
\newblock In \emph{Proceedings of the IEEE/CVF Conference on Computer Vision
  and Pattern Recognition}, pages 9665--9675, 2023.

\bibitem[Di~Palo et~al.(2024)Di~Palo, Hasenclever, Humplik, and
  Byravan]{di2024diffusion}
N.~Di~Palo, L.~Hasenclever, J.~Humplik, and A.~Byravan.
\newblock Diffusion augmented agents: A framework for efficient exploration and
  transfer learning.
\newblock \emph{arXiv preprint arXiv:2407.20798}, 2024.

\bibitem[Dimitropoulos et~al.(2022)Dimitropoulos, Hatzilygeroudis, and
  Chatzilygeroudis]{dimitropoulos2022brief}
K.~Dimitropoulos, I.~Hatzilygeroudis, and K.~Chatzilygeroudis.
\newblock A brief survey of sim2real methods for robot learning.
\newblock In \emph{International Conference on Robotics in Alpe-Adria Danube
  Region}, pages 133--140. Springer, 2022.

\bibitem[Doshi et~al.(2024)Doshi, Walke, Mees, Dasari, and
  Levine]{doshi2024scaling}
R.~Doshi, H.~Walke, O.~Mees, S.~Dasari, and S.~Levine.
\newblock Scaling cross-embodied learning: One policy for manipulation,
  navigation, locomotion and aviation.
\newblock \emph{arXiv preprint arXiv:2408.11812}, 2024.

\bibitem[Driess et~al.(2023)Driess, Xia, Sajjadi, Lynch, Chowdhery, Ichter,
  Wahid, Tompson, Vuong, Yu, et~al.]{driess2023palm}
D.~Driess, F.~Xia, M.~S. Sajjadi, C.~Lynch, A.~Chowdhery, B.~Ichter, A.~Wahid,
  J.~Tompson, Q.~Vuong, T.~Yu, et~al.
\newblock Palm-e: An embodied multimodal language model.
\newblock \emph{arXiv preprint arXiv:2303.03378}, 2023.

\bibitem[Duan et~al.(2022)Duan, Yu, Tan, Zhu, and Tan]{duan2022survey}
J.~Duan, S.~Yu, H.~L. Tan, H.~Zhu, and C.~Tan.
\newblock A survey of embodied ai: From simulators to research tasks.
\newblock \emph{IEEE Transactions on Emerging Topics in Computational
  Intelligence}, 6\penalty0 (2):\penalty0 230--244, 2022.

\bibitem[Dubey et~al.(2024)Dubey, Jauhri, Pandey, Kadian, Al-Dahle, Letman,
  Mathur, Schelten, Yang, Fan, et~al.]{dubey2024llama}
A.~Dubey, A.~Jauhri, A.~Pandey, A.~Kadian, A.~Al-Dahle, A.~Letman, A.~Mathur,
  A.~Schelten, A.~Yang, A.~Fan, et~al.
\newblock The llama 3 herd of models.
\newblock \emph{arXiv preprint arXiv:2407.21783}, 2024.

\bibitem[Ebert et~al.(2021)Ebert, Yang, Schmeckpeper, Bucher, Georgakis,
  Daniilidis, Finn, and Levine]{ebert2021bridge}
F.~Ebert, Y.~Yang, K.~Schmeckpeper, B.~Bucher, G.~Georgakis, K.~Daniilidis,
  C.~Finn, and S.~Levine.
\newblock Bridge data: Boosting generalization of robotic skills with
  cross-domain datasets.
\newblock \emph{arXiv preprint arXiv:2109.13396}, 2021.

\bibitem[Ehsani et~al.(2023)Ehsani, Gupta, Hendrix, Salvador, Weihs, Zeng,
  Singh, Kim, Han, Herrasti, et~al.]{ehsani2023imitating}
K.~Ehsani, T.~Gupta, R.~Hendrix, J.~Salvador, L.~Weihs, K.-H. Zeng, K.~P.
  Singh, Y.~Kim, W.~Han, A.~Herrasti, et~al.
\newblock Imitating shortest paths in simulation enables effective navigation
  and manipulation in the real world.
\newblock \emph{arXiv preprint arXiv:2312.02976}, 2023.

\bibitem[Ehsani et~al.(2024)Ehsani, Gupta, Hendrix, Salvador, Weihs, Zeng,
  Singh, Kim, Han, Herrasti, Krishna, Schwenk, VanderBilt, and
  Kembhavi]{ehsani2024spoc}
K.~Ehsani, T.~Gupta, R.~Hendrix, J.~Salvador, L.~Weihs, K.~Zeng, K.~Singh,
  Y.~Kim, W.~Han, A.~Herrasti, R.~Krishna, D.~Schwenk, E.~VanderBilt, and
  A.~Kembhavi.
\newblock Spoc: Imitating shortest paths in simulation enables effective
  navigation and manipulation in the real world.
\newblock In \emph{2024 IEEE/CVF Conference on Computer Vision and Pattern
  Recognition (CVPR)}, pages 16238--16250, Los Alamitos, CA, USA, jun 2024.
  IEEE Computer Society.
\newblock \doi{10.1109/CVPR52733.2024.01537}.
\newblock URL
  \url{https://doi.ieeecomputersociety.org/10.1109/CVPR52733.2024.01537}.

\bibitem[Freeman et~al.(2021)Freeman, Frey, Raichuk, Girgin, Mordatch, and
  Bachem]{freeman2021brax}
C.~D. Freeman, E.~Frey, A.~Raichuk, S.~Girgin, I.~Mordatch, and O.~Bachem.
\newblock Brax--a differentiable physics engine for large scale rigid body
  simulation.
\newblock \emph{arXiv preprint arXiv:2106.13281}, 2021.

\bibitem[Fu et~al.(2023)Fu, Song, Wu, Yu, and Scaramuzza]{song2023racing}
J.~Fu, Y.~Song, Y.~Wu, F.~Yu, and D.~Scaramuzza.
\newblock Learning deep sensorimotor policies for vision-based autonomous drone
  racing.
\newblock In \emph{2023 IEEE/RSJ International Conference on Intelligent Robots
  and Systems (IROS)}, pages 5243--5250, 2023.
\newblock \doi{10.1109/IROS55552.2023.10341805}.

\bibitem[Gan et~al.(2020)Gan, Schwartz, Alter, Mrowca, Schrimpf, Traer,
  De~Freitas, Kubilius, Bhandwaldar, Haber, et~al.]{gan2020threedworld}
C.~Gan, J.~Schwartz, S.~Alter, D.~Mrowca, M.~Schrimpf, J.~Traer, J.~De~Freitas,
  J.~Kubilius, A.~Bhandwaldar, N.~Haber, et~al.
\newblock Threedworld: A platform for interactive multi-modal physical
  simulation.
\newblock \emph{arXiv preprint arXiv:2007.04954}, 2020.

\bibitem[Gervet et~al.(2023)Gervet, Chintala, Batra, Malik, and
  Chaplot]{gervet2023navigating}
T.~Gervet, S.~Chintala, D.~Batra, J.~Malik, and D.~S. Chaplot.
\newblock Navigating to objects in the real world.
\newblock \emph{Science Robotics}, 8\penalty0 (79):\penalty0 eadf6991, 2023.

\bibitem[Gu et~al.(2023)Gu, Xiang, Li, Ling, Liu, Mu, Tang, Tao, Wei, Yao,
  et~al.]{gu2023maniskill2}
J.~Gu, F.~Xiang, X.~Li, Z.~Ling, X.~Liu, T.~Mu, Y.~Tang, S.~Tao, X.~Wei,
  Y.~Yao, et~al.
\newblock Maniskill2: A unified benchmark for generalizable manipulation
  skills.
\newblock \emph{arXiv preprint arXiv:2302.04659}, 2023.

\bibitem[Haarnoja et~al.(2024)Haarnoja, Moran, Lever, Huang, Tirumala, Humplik,
  Wulfmeier, Tunyasuvunakool, Siegel, Hafner, et~al.]{haarnoja2024learning}
T.~Haarnoja, B.~Moran, G.~Lever, S.~H. Huang, D.~Tirumala, J.~Humplik,
  M.~Wulfmeier, S.~Tunyasuvunakool, N.~Y. Siegel, R.~Hafner, et~al.
\newblock Learning agile soccer skills for a bipedal robot with deep
  reinforcement learning.
\newblock \emph{Science Robotics}, 9\penalty0 (89):\penalty0 eadi8022, 2024.

\bibitem[Hu et~al.(2024)Hu, Hendrix, Farhadi, Kembhavi, Martin-Martin, Stone,
  Zeng, and Ehsan]{hu2024flare}
J.~Hu, R.~Hendrix, A.~Farhadi, A.~Kembhavi, R.~Martin-Martin, P.~Stone, K.-H.
  Zeng, and K.~Ehsan.
\newblock Flare: Achieving masterful and adaptive robot policies with
  large-scale reinforcement learning fine-tuning.
\newblock \emph{arXiv preprint arXiv:2409.16578}, 2024.

\bibitem[Hua et~al.(2024)Hua, Liu, Macaluso, Lin, Zhang, Xu, and
  Wang]{gensim2024}
P.~Hua, M.~Liu, A.~Macaluso, Y.~Lin, W.~Zhang, H.~Xu, and L.~Wang.
\newblock {GenSim2: Scaling robot data generation with multi-modal and
  reasoning LLMs}.
\newblock In \emph{8th Annual Conference on Robot Learning (CoRL)}, 2024.

\bibitem[Hwangbo et~al.(2019)Hwangbo, Lee, Dosovitskiy, Bellicoso, Tsounis,
  Koltun, and Hutter]{hwangbo2019learning}
J.~Hwangbo, J.~Lee, A.~Dosovitskiy, D.~Bellicoso, V.~Tsounis, V.~Koltun, and
  M.~Hutter.
\newblock Learning agile and dynamic motor skills for legged robots.
\newblock \emph{Science Robotics}, 4\penalty0 (26), 2019.

\bibitem[Khandelwal et~al.(2022)Khandelwal, Weihs, Mottaghi, and
  Kembhavi]{khandelwal2022simple}
A.~Khandelwal, L.~Weihs, R.~Mottaghi, and A.~Kembhavi.
\newblock Simple but effective: Clip embeddings for embodied ai.
\newblock In \emph{Proceedings of the IEEE/CVF Conference on Computer Vision
  and Pattern Recognition}, pages 14829--14838, 2022.

\bibitem[Kim et~al.(2024)Kim, Pertsch, Karamcheti, Xiao, Balakrishna, Nair,
  Rafailov, Foster, Lam, Sanketi, et~al.]{kim2024openvla}
M.~J. Kim, K.~Pertsch, S.~Karamcheti, T.~Xiao, A.~Balakrishna, S.~Nair,
  R.~Rafailov, E.~Foster, G.~Lam, P.~Sanketi, et~al.
\newblock Openvla: An open-source vision-language-action model.
\newblock \emph{arXiv preprint arXiv:2406.09246}, 2024.

\bibitem[Kolve et~al.(2017)Kolve, Mottaghi, Han, VanderBilt, Weihs, Herrasti,
  Deitke, Ehsani, Gordon, Zhu, et~al.]{kolve2017ai2}
E.~Kolve, R.~Mottaghi, W.~Han, E.~VanderBilt, L.~Weihs, A.~Herrasti, M.~Deitke,
  K.~Ehsani, D.~Gordon, Y.~Zhu, et~al.
\newblock Ai2-thor: An interactive 3d environment for visual ai.
\newblock \emph{arXiv preprint arXiv:1712.05474}, 2017.

\bibitem[Kumar et~al.(2021)Kumar, Fu, Pathak, and Malik]{kumar2021rma}
A.~Kumar, Z.~Fu, D.~Pathak, and J.~Malik.
\newblock Rma: Rapid motor adaptation for legged robots.
\newblock \emph{arXiv preprint arXiv:2107.04034}, 2021.

\bibitem[Lee et~al.(2021)Lee, Devin, Zhou, Lampe, Bousmalis, Springenberg,
  Byravan, Abdolmaleki, Gileadi, Khosid, et~al.]{lee2021beyond}
A.~X. Lee, C.~M. Devin, Y.~Zhou, T.~Lampe, K.~Bousmalis, J.~T. Springenberg,
  A.~Byravan, A.~Abdolmaleki, N.~Gileadi, D.~Khosid, et~al.
\newblock Beyond pick-and-place: Tackling robotic stacking of diverse shapes.
\newblock In \emph{5th Annual Conference on Robot Learning}, 2021.

\bibitem[Lee et~al.(2020{\natexlab{a}})Lee, Hwangbo, Wellhausen, Koltun, and
  Hutter]{Lee_2020}
J.~Lee, J.~Hwangbo, L.~Wellhausen, V.~Koltun, and M.~Hutter.
\newblock Learning quadrupedal locomotion over challenging terrain.
\newblock \emph{Science Robotics}, 5\penalty0 (47):\penalty0 eabc5986, Oct
  2020{\natexlab{a}}.
\newblock ISSN 2470-9476.
\newblock \doi{10.1126/scirobotics.abc5986}.
\newblock URL \url{http://dx.doi.org/10.1126/scirobotics.abc5986}.

\bibitem[Lee et~al.(2020{\natexlab{b}})Lee, Hwangbo, Wellhausen, Koltun, and
  Hutter]{lee2020challenging}
J.~Lee, J.~Hwangbo, L.~Wellhausen, V.~Koltun, and M.~Hutter.
\newblock Learning quadrupedal locomotion over challenging terrain.
\newblock \emph{Science Robotics}, 5\penalty0 (47):\penalty0 eabc5986,
  2020{\natexlab{b}}.
\newblock \doi{10.1126/scirobotics.abc5986}.
\newblock URL
  \url{https://www.science.org/doi/abs/10.1126/scirobotics.abc5986}.

\bibitem[Li et~al.(2023)Li, Zhang, Wong, Gokmen, Srivastava,
  Mart{\'\i}n-Mart{\'\i}n, Wang, Levine, Lingelbach, Sun,
  et~al.]{li2023behavior}
C.~Li, R.~Zhang, J.~Wong, C.~Gokmen, S.~Srivastava, R.~Mart{\'\i}n-Mart{\'\i}n,
  C.~Wang, G.~Levine, M.~Lingelbach, J.~Sun, et~al.
\newblock Behavior-1k: A benchmark for embodied ai with 1,000 everyday
  activities and realistic simulation.
\newblock In \emph{Conference on Robot Learning}, pages 80--93. PMLR, 2023.

\bibitem[Li et~al.(2021)Li, Cheng, Peng, Abbeel, Levine, Berseth, and
  Sreenath]{li2021reinforcement}
Z.~Li, X.~Cheng, X.~B. Peng, P.~Abbeel, S.~Levine, G.~Berseth, and K.~Sreenath.
\newblock Reinforcement learning for robust parameterized locomotion control of
  bipedal robots.
\newblock \emph{arXiv preprint arXiv:2103.14295}, 2021.

\bibitem[Liang et~al.(2023)Liang, Huang, Xia, Xu, Hausman, Ichter, Florence,
  and Zeng]{liang2023code}
J.~Liang, W.~Huang, F.~Xia, P.~Xu, K.~Hausman, B.~Ichter, P.~Florence, and
  A.~Zeng.
\newblock Code as policies: Language model programs for embodied control.
\newblock In \emph{2023 IEEE International Conference on Robotics and
  Automation (ICRA)}, pages 9493--9500. IEEE, 2023.

\bibitem[Lu et~al.(2024)Lu, Clark, Lee, Zhang, Khosla, Marten, Hoiem, and
  Kembhavi]{lu2024unified}
J.~Lu, C.~Clark, S.~Lee, Z.~Zhang, S.~Khosla, R.~Marten, D.~Hoiem, and
  A.~Kembhavi.
\newblock Unified-io 2: Scaling autoregressive multimodal models with vision
  language audio and action.
\newblock In \emph{Proceedings of the IEEE/CVF Conference on Computer Vision
  and Pattern Recognition}, pages 26439--26455, 2024.

\bibitem[Ma et~al.(2023)Ma, Liang, Wang, Huang, Bastani, Jayaraman, Zhu, Fan,
  and Anandkumar]{ma2023eureka}
Y.~J. Ma, W.~Liang, G.~Wang, D.-A. Huang, O.~Bastani, D.~Jayaraman, Y.~Zhu,
  L.~Fan, and A.~Anandkumar.
\newblock Eureka: Human-level reward design via coding large language models.
\newblock \emph{arXiv preprint arXiv:2310.12931}, 2023.

\bibitem[Majumdar et~al.(2023)Majumdar, Yadav, Arnaud, Ma, Chen, Silwal, Jain,
  Berges, Wu, Vakil, et~al.]{majumdar2023we}
A.~Majumdar, K.~Yadav, S.~Arnaud, J.~Ma, C.~Chen, S.~Silwal, A.~Jain, V.-P.
  Berges, T.~Wu, J.~Vakil, et~al.
\newblock Where are we in the search for an artificial visual cortex for
  embodied intelligence?
\newblock \emph{Advances in Neural Information Processing Systems},
  36:\penalty0 655--677, 2023.

\bibitem[Makoviychuk et~al.(2021)Makoviychuk, Wawrzyniak, Guo, Lu, Storey,
  Macklin, Hoeller, Rudin, Allshire, Handa, et~al.]{makoviychuk2021isaac}
V.~Makoviychuk, L.~Wawrzyniak, Y.~Guo, M.~Lu, K.~Storey, M.~Macklin,
  D.~Hoeller, N.~Rudin, A.~Allshire, A.~Handa, et~al.
\newblock Isaac gym: High performance gpu-based physics simulation for robot
  learning.
\newblock \emph{arXiv preprint arXiv:2108.10470}, 2021.

\bibitem[Margolis et~al.(2022)Margolis, Chen, Paigwar, Fu, Kim, Kim, and
  Agrawal]{margolis2022jump}
G.~B. Margolis, T.~Chen, K.~Paigwar, X.~Fu, D.~Kim, S.~b. Kim, and P.~Agrawal.
\newblock Learning to jump from pixels.
\newblock In A.~Faust, D.~Hsu, and G.~Neumann, editors, \emph{Proceedings of
  the 5th Conference on Robot Learning}, volume 164 of \emph{Proceedings of
  Machine Learning Research}, pages 1025--1034. PMLR, 08--11 Nov 2022.
\newblock URL \url{https://proceedings.mlr.press/v164/margolis22a.html}.

\bibitem[Miki et~al.(2022)Miki, Lee, Hwangbo, Wellhausen, Koltun, and
  Hutter]{miki2022perceptive}
T.~Miki, J.~Lee, J.~Hwangbo, L.~Wellhausen, V.~Koltun, and M.~Hutter.
\newblock Learning robust perceptive locomotion for quadrupedal robots in the
  wild.
\newblock \emph{Science Robotics}, 7\penalty0 (62):\penalty0 eabk2822, 2022.
\newblock \doi{10.1126/scirobotics.abk2822}.
\newblock URL
  \url{https://www.science.org/doi/abs/10.1126/scirobotics.abk2822}.

\bibitem[Mu et~al.(2021)Mu, Ling, Xiang, Yang, Li, Tao, Huang, Jia, and
  Su]{mu2021maniskill}
T.~Mu, Z.~Ling, F.~Xiang, D.~Yang, X.~Li, S.~Tao, Z.~Huang, Z.~Jia, and H.~Su.
\newblock Maniskill: Generalizable manipulation skill benchmark with
  large-scale demonstrations.
\newblock \emph{arXiv preprint arXiv:2107.14483}, 2021.

\bibitem[Muratore et~al.(2022)Muratore, Ramos, Turk, Yu, Gienger, and
  Peters]{muratore2022robot}
F.~Muratore, F.~Ramos, G.~Turk, W.~Yu, M.~Gienger, and J.~Peters.
\newblock Robot learning from randomized simulations: A review.
\newblock \emph{Frontiers in Robotics and AI}, 9:\penalty0 799893, 2022.

\bibitem[Nasiriany et~al.(2024{\natexlab{a}})Nasiriany, Maddukuri, Zhang,
  Parikh, Lo, Joshi, Mandlekar, and Zhu]{nasiriany2024robocasa}
S.~Nasiriany, A.~Maddukuri, L.~Zhang, A.~Parikh, A.~Lo, A.~Joshi, A.~Mandlekar,
  and Y.~Zhu.
\newblock Robocasa: Large-scale simulation of everyday tasks for generalist
  robots.
\newblock \emph{arXiv preprint arXiv:2406.02523}, 2024{\natexlab{a}}.

\bibitem[Nasiriany et~al.(2024{\natexlab{b}})Nasiriany, Maddukuri, Zhang,
  Parikh, Lo, Joshi, Mandlekar, and Zhu]{robocasa2024}
S.~Nasiriany, A.~Maddukuri, L.~Zhang, A.~Parikh, A.~Lo, A.~Joshi, A.~Mandlekar,
  and Y.~Zhu.
\newblock Robocasa: Large-scale simulation of everyday tasks for generalist
  robots.
\newblock In \emph{Robotics: Science and Systems}, 2024{\natexlab{b}}.

\bibitem[Ouyang et~al.(2024)Ouyang, Li, Li, Li, Yu, Sreenath, and
  Wu]{ouyang2024long}
Y.~Ouyang, J.~Li, Y.~Li, Z.~Li, C.~Yu, K.~Sreenath, and Y.~Wu.
\newblock Long-horizon locomotion and manipulation on a quadrupedal robot with
  large language models.
\newblock \emph{arXiv preprint arXiv:2404.05291}, 2024.

\bibitem[Padalkar et~al.(2023)Padalkar, Pooley, Jain, Bewley, Herzog, Irpan,
  Khazatsky, Rai, Singh, Brohan, et~al.]{padalkar2023open}
A.~Padalkar, A.~Pooley, A.~Jain, A.~Bewley, A.~Herzog, A.~Irpan, A.~Khazatsky,
  A.~Rai, A.~Singh, A.~Brohan, et~al.
\newblock Open x-embodiment: Robotic learning datasets and rt-x models.
\newblock \emph{arXiv preprint arXiv:2310.08864}, 2023.

\bibitem[Peng et~al.(2020)Peng, Coumans, Zhang, Lee, Tan, and
  Levine]{peng2020learning}
X.~B. Peng, E.~Coumans, T.~Zhang, T.-W. Lee, J.~Tan, and S.~Levine.
\newblock Learning agile robotic locomotion skills by imitating animals.
\newblock \emph{arXiv preprint arXiv:2004.00784}, 2020.

\bibitem[Puig et~al.(2023)Puig, Undersander, Szot, Cote, Yang, Partsey, Desai,
  Clegg, Hlavac, Min, et~al.]{puig2023habitat}
X.~Puig, E.~Undersander, A.~Szot, M.~D. Cote, T.-Y. Yang, R.~Partsey, R.~Desai,
  A.~W. Clegg, M.~Hlavac, S.~Y. Min, et~al.
\newblock Habitat 3.0: A co-habitat for humans, avatars and robots.
\newblock \emph{arXiv preprint arXiv:2310.13724}, 2023.

\bibitem[Rajeswaran et~al.(2017)Rajeswaran, Kumar, Gupta, Vezzani, Schulman,
  Todorov, and Levine]{rajeswaran2017learning}
A.~Rajeswaran, V.~Kumar, A.~Gupta, G.~Vezzani, J.~Schulman, E.~Todorov, and
  S.~Levine.
\newblock Learning complex dexterous manipulation with deep reinforcement
  learning and demonstrations.
\newblock \emph{arXiv preprint arXiv:1709.10087}, 2017.

\bibitem[Reed et~al.(2022)Reed, Zolna, Parisotto, Colmenarejo, Novikov,
  Barth-Maron, Gimenez, Sulsky, Kay, Springenberg, et~al.]{reed2022generalist}
S.~Reed, K.~Zolna, E.~Parisotto, S.~G. Colmenarejo, A.~Novikov, G.~Barth-Maron,
  M.~Gimenez, Y.~Sulsky, J.~Kay, J.~T. Springenberg, et~al.
\newblock A generalist agent.
\newblock \emph{arXiv preprint arXiv:2205.06175}, 2022.

\bibitem[Reid et~al.(2024)Reid, Savinov, Teplyashin, Lepikhin, Lillicrap,
  Alayrac, Soricut, Lazaridou, Firat, Schrittwieser, et~al.]{reid2024gemini}
M.~Reid, N.~Savinov, D.~Teplyashin, D.~Lepikhin, T.~Lillicrap, J.-b. Alayrac,
  R.~Soricut, A.~Lazaridou, O.~Firat, J.~Schrittwieser, et~al.
\newblock Gemini 1.5: Unlocking multimodal understanding across millions of
  tokens of context.
\newblock \emph{arXiv preprint arXiv:2403.05530}, 2024.

\bibitem[Sadeghi and Levine(2016)]{sadeghi2016cad2rl}
F.~Sadeghi and S.~Levine.
\newblock Cad2rl: Real single-image flight without a single real image.
\newblock \emph{arXiv preprint arXiv:1611.04201}, 2016.

\bibitem[Savva et~al.(2019)Savva, Kadian, Maksymets, Zhao, Wijmans, Jain,
  Straub, Liu, Koltun, Malik, et~al.]{savva2019habitat}
M.~Savva, A.~Kadian, O.~Maksymets, Y.~Zhao, E.~Wijmans, B.~Jain, J.~Straub,
  J.~Liu, V.~Koltun, J.~Malik, et~al.
\newblock Habitat: A platform for embodied ai research.
\newblock In \emph{Proceedings of the IEEE/CVF international conference on
  computer vision}, pages 9339--9347, 2019.

\bibitem[Scholz et~al.(2011)Scholz, Chitta, Marthi, and
  Likhachev]{scholz2011cart}
J.~Scholz, S.~Chitta, B.~Marthi, and M.~Likhachev.
\newblock Cart pushing with a mobile manipulation system: Towards navigation
  with moveable objects.
\newblock In \emph{2011 IEEE International Conference on Robotics and
  Automation}, pages 6115--6120. IEEE, 2011.

\bibitem[Shafiullah et~al.(2022)Shafiullah, Paxton, Pinto, Chintala, and
  Szlam]{shafiullah2022clip}
N.~M.~M. Shafiullah, C.~Paxton, L.~Pinto, S.~Chintala, and A.~Szlam.
\newblock Clip-fields: Weakly supervised semantic fields for robotic memory.
\newblock \emph{arXiv preprint arXiv:2210.05663}, 2022.

\bibitem[Shah et~al.(2023{\natexlab{a}})Shah, Osi{\'n}ski, Levine,
  et~al.]{shah2023lm}
D.~Shah, B.~Osi{\'n}ski, S.~Levine, et~al.
\newblock Lm-nav: Robotic navigation with large pre-trained models of language,
  vision, and action.
\newblock In \emph{Conference on robot learning}, pages 492--504. PMLR,
  2023{\natexlab{a}}.

\bibitem[Shah et~al.(2023{\natexlab{b}})Shah, Sridhar, Bhorkar, Hirose, and
  Levine]{shah2023gnm}
D.~Shah, A.~Sridhar, A.~Bhorkar, N.~Hirose, and S.~Levine.
\newblock Gnm: A general navigation model to drive any robot.
\newblock In \emph{2023 IEEE International Conference on Robotics and
  Automation (ICRA)}, pages 7226--7233. IEEE, 2023{\natexlab{b}}.

\bibitem[Shah et~al.(2023{\natexlab{c}})Shah, Sridhar, Dashora, Stachowicz,
  Black, Hirose, and Levine]{shah2023vint}
D.~Shah, A.~Sridhar, N.~Dashora, K.~Stachowicz, K.~Black, N.~Hirose, and
  S.~Levine.
\newblock Vint: A foundation model for visual navigation.
\newblock \emph{arXiv preprint arXiv:2306.14846}, 2023{\natexlab{c}}.

\bibitem[Shen et~al.(2021)Shen, Xia, Li, Mart{\'\i}n-Mart{\'\i}n, Fan, Wang,
  P{\'e}rez-D’Arpino, Buch, Srivastava, Tchapmi, et~al.]{shen2021igibson}
B.~Shen, F.~Xia, C.~Li, R.~Mart{\'\i}n-Mart{\'\i}n, L.~Fan, G.~Wang,
  C.~P{\'e}rez-D’Arpino, S.~Buch, S.~Srivastava, L.~Tchapmi, et~al.
\newblock igibson 1.0: a simulation environment for interactive tasks in large
  realistic scenes.
\newblock In \emph{2021 IEEE/RSJ International Conference on Intelligent Robots
  and Systems (IROS)}, pages 7520--7527. IEEE, 2021.

\bibitem[Siekmann et~al.(2021)Siekmann, Green, Warila, Fern, and
  Hurst]{siekmann2021blind}
J.~Siekmann, K.~Green, J.~Warila, A.~Fern, and J.~Hurst.
\newblock Blind bipedal stair traversal via sim-to-real reinforcement learning.
\newblock \emph{arXiv preprint arXiv:2105.08328}, 2021.

\bibitem[Song et~al.(2023)Song, Shi, Penicka, and
  Scaramuzza]{song2023cluttered}
Y.~Song, K.~Shi, R.~Penicka, and D.~Scaramuzza.
\newblock Learning perception-aware agile flight in cluttered environments.
\newblock In \emph{2023 IEEE International Conference on Robotics and
  Automation (ICRA)}, pages 1989--1995, 2023.
\newblock \doi{10.1109/ICRA48891.2023.10160563}.

\bibitem[Szot et~al.(2021)Szot, Clegg, Undersander, Wijmans, Zhao, Turner,
  Maestre, Mukadam, Chaplot, Maksymets, et~al.]{szot2021habitat}
A.~Szot, A.~Clegg, E.~Undersander, E.~Wijmans, Y.~Zhao, J.~Turner, N.~Maestre,
  M.~Mukadam, D.~S. Chaplot, O.~Maksymets, et~al.
\newblock Habitat 2.0: Training home assistants to rearrange their habitat.
\newblock \emph{Advances in neural information processing systems},
  34:\penalty0 251--266, 2021.

\bibitem[Team et~al.(2023)Team, Anil, Borgeaud, Wu, Alayrac, Yu, Soricut,
  Schalkwyk, Dai, Hauth, et~al.]{team2023gemini}
G.~Team, R.~Anil, S.~Borgeaud, Y.~Wu, J.-B. Alayrac, J.~Yu, R.~Soricut,
  J.~Schalkwyk, A.~M. Dai, A.~Hauth, et~al.
\newblock Gemini: a family of highly capable multimodal models.
\newblock \emph{arXiv preprint arXiv:2312.11805}, 2023.

\bibitem[Team(2025)]{gemini_for_robotics}
G.~R. Team.
\newblock {Gemini Robotics: Bringing AI into the physical world}.
\newblock \emph{arXiv preprint}, March 2025.

\bibitem[Team et~al.(2024)Team, Ghosh, Walke, Pertsch, Black, Mees, Dasari,
  Hejna, Kreiman, Xu, et~al.]{team2024octo}
O.~M. Team, D.~Ghosh, H.~Walke, K.~Pertsch, K.~Black, O.~Mees, S.~Dasari,
  J.~Hejna, T.~Kreiman, C.~Xu, et~al.
\newblock Octo: An open-source generalist robot policy.
\newblock \emph{arXiv preprint arXiv:2405.12213}, 2024.

\bibitem[Tobin et~al.(2017)Tobin, Fong, Ray, Schneider, Zaremba, and
  Abbeel]{tobin2017domain}
J.~Tobin, R.~Fong, A.~Ray, J.~Schneider, W.~Zaremba, and P.~Abbeel.
\newblock Domain randomization for transferring deep neural networks from
  simulation to the real world.
\newblock In \emph{2017 IEEE/RSJ international conference on intelligent robots
  and systems (IROS)}, pages 23--30. IEEE, 2017.

\bibitem[Todorov et~al.(2012)Todorov, Erez, and Tassa]{todorov2012mujoco}
E.~Todorov, T.~Erez, and Y.~Tassa.
\newblock Mujoco: A physics engine for model-based control.
\newblock In \emph{2012 IEEE/RSJ International Conference on Intelligent Robots
  and Systems}, pages 5026--5033. IEEE, 2012.
\newblock \doi{10.1109/IROS.2012.6386109}.

\bibitem[{Unity Technologies}(2023)]{unity}
{Unity Technologies}.
\newblock Unity, 2023.
\newblock URL \url{https://unity.com/}.
\newblock Game development platform.

\bibitem[Vasilopoulos et~al.(2018)Vasilopoulos, Topping, Vega-Brown, Roy, and
  Koditschek]{vasilopoulos2018sensor}
V.~Vasilopoulos, T.~T. Topping, W.~Vega-Brown, N.~Roy, and D.~E. Koditschek.
\newblock Sensor-based reactive execution of symbolic rearrangement plans by a
  legged mobile manipulator.
\newblock In \emph{2018 IEEE/RSJ International Conference on Intelligent Robots
  and Systems (IROS)}, pages 3298--3305. IEEE, 2018.

\bibitem[Wang et~al.(2023{\natexlab{a}})Wang, Ling, Yuan, Shridhar, Bao, Qin,
  Wang, Xu, and Wang]{wang2023gensim}
L.~Wang, Y.~Ling, Z.~Yuan, M.~Shridhar, C.~Bao, Y.~Qin, B.~Wang, H.~Xu, and
  X.~Wang.
\newblock Gensim: Generating robotic simulation tasks via large language
  models.
\newblock \emph{arXiv preprint arXiv:2310.01361}, 2023{\natexlab{a}}.

\bibitem[Wang et~al.(2023{\natexlab{b}})Wang, Xian, Chen, Wang, Wang,
  Fragkiadaki, Erickson, Held, and Gan]{wang2023robogen}
Y.~Wang, Z.~Xian, F.~Chen, T.-H. Wang, Y.~Wang, K.~Fragkiadaki, Z.~Erickson,
  D.~Held, and C.~Gan.
\newblock Robogen: Towards unleashing infinite data for automated robot
  learning via generative simulation.
\newblock \emph{arXiv preprint arXiv:2311.01455}, 2023{\natexlab{b}}.

\bibitem[Wijmans et~al.(2019)Wijmans, Kadian, Morcos, Lee, Essa, Parikh, Savva,
  and Batra]{wijmans2019dd}
E.~Wijmans, A.~Kadian, A.~Morcos, S.~Lee, I.~Essa, D.~Parikh, M.~Savva, and
  D.~Batra.
\newblock Dd-ppo: Learning near-perfect pointgoal navigators from 2.5 billion
  frames.
\newblock \emph{arXiv preprint arXiv:1911.00357}, 2019.

\bibitem[Xia et~al.(2018)Xia, Zamir, He, Sax, Malik, and
  Savarese]{xia2018gibson}
F.~Xia, A.~R. Zamir, Z.~He, A.~Sax, J.~Malik, and S.~Savarese.
\newblock Gibson env: Real-world perception for embodied agents.
\newblock In \emph{Proceedings of the IEEE conference on computer vision and
  pattern recognition}, pages 9068--9079, 2018.

\bibitem[Xia et~al.(2019)Xia, Li, Chen, Shen, Mart{\i}n-Mart{\i}n, Hirose,
  Zamir, Fei-Fei, and Savarese]{xia2019gibson}
F.~Xia, C.~Li, K.~Chen, W.~B. Shen, R.~Mart{\i}n-Mart{\i}n, N.~Hirose, A.~R.
  Zamir, L.~Fei-Fei, and S.~Savarese.
\newblock Gibson env v2: Embodied simulation environments for interactive
  navigation.
\newblock \emph{Stanford University, Tech. Rep.}, 2019.

\bibitem[Xiang et~al.(2020)Xiang, Qin, Mo, Xia, Zhu, Liu, Liu, Jiang, Yuan,
  Wang, et~al.]{xiang2020sapien}
F.~Xiang, Y.~Qin, K.~Mo, Y.~Xia, H.~Zhu, F.~Liu, M.~Liu, H.~Jiang, Y.~Yuan,
  H.~Wang, et~al.
\newblock Sapien: A simulated part-based interactive environment.
\newblock In \emph{Proceedings of the IEEE/CVF conference on computer vision
  and pattern recognition}, pages 11097--11107, 2020.

\bibitem[Yang et~al.(2024{\natexlab{a}})Yang, Glossop, Bhorkar, Shah, Vuong,
  Finn, Sadigh, and Levine]{yang2024pushing}
J.~Yang, C.~Glossop, A.~Bhorkar, D.~Shah, Q.~Vuong, C.~Finn, D.~Sadigh, and
  S.~Levine.
\newblock Pushing the limits of cross-embodiment learning for manipulation and
  navigation.
\newblock \emph{arXiv preprint arXiv:2402.19432}, 2024{\natexlab{a}}.

\bibitem[Yang et~al.(2024{\natexlab{b}})Yang, Sun, Weihs, VanderBilt, Herrasti,
  Han, Wu, Haber, Krishna, Liu, et~al.]{yang2024holodeck}
Y.~Yang, F.-Y. Sun, L.~Weihs, E.~VanderBilt, A.~Herrasti, W.~Han, J.~Wu,
  N.~Haber, R.~Krishna, L.~Liu, et~al.
\newblock Holodeck: Language guided generation of 3d embodied ai environments.
\newblock In \emph{Proceedings of the IEEE/CVF Conference on Computer Vision
  and Pattern Recognition}, pages 16227--16237, 2024{\natexlab{b}}.

\bibitem[Yenamandra et~al.(2023)Yenamandra, Ramachandran, Yadav, Wang, Khanna,
  Gervet, Yang, Jain, Clegg, Turner, et~al.]{yenamandra2023homerobot}
S.~Yenamandra, A.~Ramachandran, K.~Yadav, A.~Wang, M.~Khanna, T.~Gervet, T.-Y.
  Yang, V.~Jain, A.~W. Clegg, J.~Turner, et~al.
\newblock Homerobot: Open-vocabulary mobile manipulation.
\newblock \emph{arXiv preprint arXiv:2306.11565}, 2023.

\bibitem[Yu et~al.(2023{\natexlab{a}})Yu, Xiao, Stone, Tompson, Brohan, Wang,
  Singh, Tan, Peralta, Ichter, et~al.]{yu2023scaling}
T.~Yu, T.~Xiao, A.~Stone, J.~Tompson, A.~Brohan, S.~Wang, J.~Singh, C.~Tan,
  J.~Peralta, B.~Ichter, et~al.
\newblock Scaling robot learning with semantically imagined experience.
\newblock \emph{arXiv preprint arXiv:2302.11550}, 2023{\natexlab{a}}.

\bibitem[Yu et~al.(2019)Yu, Kumar, Turk, and Liu]{yu2019sim}
W.~Yu, V.~C. Kumar, G.~Turk, and C.~K. Liu.
\newblock Sim-to-real transfer for biped locomotion.
\newblock \emph{arXiv preprint arXiv:1903.01390}, 2019.

\bibitem[Yu et~al.(2023{\natexlab{b}})Yu, Gileadi, Fu, Kirmani, Lee, Arenas,
  Chiang, Erez, Hasenclever, Humplik, et~al.]{yu2023language}
W.~Yu, N.~Gileadi, C.~Fu, S.~Kirmani, K.-H. Lee, M.~G. Arenas, H.-T.~L. Chiang,
  T.~Erez, L.~Hasenclever, J.~Humplik, et~al.
\newblock Language to rewards for robotic skill synthesis.
\newblock \emph{arXiv preprint arXiv:2306.08647}, 2023{\natexlab{b}}.

\bibitem[Zakka et~al.(2022)Zakka, Tassa, and {MuJoCo Menagerie
  Contributors}]{menagerie2022github}
K.~Zakka, Y.~Tassa, and {MuJoCo Menagerie Contributors}.
\newblock {MuJoCo Menagerie: A collection of high-quality simulation models for
  MuJoCo}, 2022.
\newblock URL \url{http://github.com/google-deepmind/mujoco_menagerie}.

\bibitem[Zeng et~al.(2022)Zeng, Attarian, Ichter, Choromanski, Wong, Welker,
  Tombari, Purohit, Ryoo, Sindhwani, et~al.]{zeng2022socratic}
A.~Zeng, M.~Attarian, B.~Ichter, K.~Choromanski, A.~Wong, S.~Welker,
  F.~Tombari, A.~Purohit, M.~Ryoo, V.~Sindhwani, et~al.
\newblock Socratic models: Composing zero-shot multimodal reasoning with
  language.
\newblock \emph{arXiv preprint arXiv:2204.00598}, 2022.

\bibitem[Zeng et~al.(2024)Zeng, Zhang, Ehsani, Hendrix, Salvador, Herrasti,
  Girshick, Kembhavi, and Weihs]{zeng2024poliformer}
K.-H. Zeng, Z.~Zhang, K.~Ehsani, R.~Hendrix, J.~Salvador, A.~Herrasti,
  R.~Girshick, A.~Kembhavi, and L.~Weihs.
\newblock Poliformer: Scaling on-policy rl with transformers results in
  masterful navigators.
\newblock \emph{arXiv preprint arXiv:2406.20083}, 2024.

\bibitem[Zhai et~al.(2023)Zhai, Mustafa, Kolesnikov, and
  Beyer]{zhai2023sigmoid}
X.~Zhai, B.~Mustafa, A.~Kolesnikov, and L.~Beyer.
\newblock Sigmoid loss for language image pre-training.
\newblock In \emph{Proceedings of the IEEE/CVF International Conference on
  Computer Vision}, pages 11975--11986, 2023.

\bibitem[Zhuang et~al.(2023)Zhuang, Fu, Wang, Atkeson, Schwertfeger, Finn, and
  Zhao]{ziwen2023parkour}
Z.~Zhuang, Z.~Fu, J.~Wang, C.~G. Atkeson, S.~Schwertfeger, C.~Finn, and
  H.~Zhao.
\newblock Robot parkour learning.
\newblock In J.~Tan, M.~Toussaint, and K.~Darvish, editors, \emph{Proceedings
  of The 7th Conference on Robot Learning}, volume 229 of \emph{Proceedings of
  Machine Learning Research}, pages 73--92. PMLR, 06--09 Nov 2023.
\newblock URL \url{https://proceedings.mlr.press/v229/zhuang23a.html}.

\end{thebibliography}
